\documentclass[journal]{IEEEtran}

\ifCLASSINFOpdf
\else
\fi
\usepackage{cite}
\usepackage{amsmath,amssymb,amsfonts}
\usepackage{algorithmic}
\usepackage{algorithm}
\usepackage{graphicx}
\usepackage{subfigure}
\usepackage{textcomp}
\usepackage{booktabs}
\usepackage{graphicx}
\usepackage{amsmath}
\usepackage{subfigure} 
\usepackage{tabularx}
\usepackage[table,xcdraw]{xcolor}
\usepackage[colorlinks,linkcolor=blue,citecolor=blue]{hyperref}
\usepackage{multirow}
\usepackage{svg}
\usepackage{float}

\usepackage{amsmath, mathtools} 

\def\BibTeX{{\rm B\kern-.05em{\sc i\kern-.025em b}\kern-.08em
    T\kern-.1667em\lower.7ex\hbox{E}\kern-.125emX}}

\begin{document}

\title{Chain-of-Thought for Autonomous Driving:\\ A Comprehensive Survey and Future Prospects}

\author{Yixin Cui, Haotian Lin, Shuo Yang, Yixiao Wang\\
Yanjun Huang, and Hong Chen,~\IEEEmembership{Fellow,~IEEE}

\thanks{Manuscript received 20 January 2025. 
This work was supported by the National Key Research and Development Program of China under Grant 2022YFB2502900 and the National Natural Science Foundation of China, Joint Fund for Innovative Enterprise Development (U23B2061). (Corresponding author: Yanjun Huang.)}
\thanks{Yixin Cui, Shuo Yang and Yixiao Wang are with the School of Automotive Studies, Tongji University, Shanghai 201804, China (e-mail: 2411448@tongji.edu.cn; 2111550@tongji.edu.cn; 2353147@tongji.edu.cn).}
\thanks{Haotian Lin is with the School of Physics Science and Engineering, Tongji University, Shanghai 200092, China (e-mail: 2250120@tongji.edu.cn).}
\thanks{Yanjun Huang is with the School of Automotive Studies, Tongji University, Shanghai 201804, China, and also with the Frontiers Science Center for Intelligent Autonomous Systems, Shanghai 200120, China (e-mail: yanjun\_huang@tongji.edu.cn).}
\thanks{Hong Chen is with the College of Electronics and Information Engineering and the Clean Energy Automotive Engineering Center, Tongji University, Shanghai 201804, China (e-mail: chenhong2019@tongji.edu.cn).}
}


\maketitle
\begin{abstract}
The rapid evolution of large language models in natural language processing has substantially elevated their semantic understanding and logical reasoning capabilities. Such proficiencies have been leveraged in autonomous driving systems, contributing to significant improvements in system performance. Models such as OpenAI o1 and DeepSeek-R1, leverage Chain-of-Thought (CoT) reasoning, an advanced cognitive method that simulates human thinking processes, demonstrating remarkable reasoning capabilities in complex tasks. By structuring complex driving scenarios within a systematic reasoning framework, this approach has emerged as a prominent research focus in autonomous driving, substantially improving the system's ability to handle challenging cases. This paper investigates how CoT methods improve the reasoning abilities of autonomous driving models. Based on a comprehensive literature review, we present a systematic analysis of the motivations, methodologies, challenges, and future research directions of CoT in autonomous driving. Furthermore, we propose the insight of combining CoT with self-learning to facilitate self-evolution in driving systems. To ensure the relevance and timeliness of this study, we have compiled a dynamic repository of literature and open-source projects, diligently updated to incorporate forefront developments. The repository is publicly available at \url{https://github.com/cuiyx1720/Awesome-CoT4AD}.

\end{abstract}

\begin{IEEEkeywords}
Chain-of-Thought, Autonomous Driving, Large Language Models, Reasoning, Self-evolution, Reflection
\end{IEEEkeywords}

\IEEEpeerreviewmaketitle

\section{Introduction}
\IEEEPARstart{T}{he} advent of large language models (LLMs) marks a significant breakthrough in the field of artificial intelligence \cite{vaswani2017attention} \cite{brown2020language}. Through pre-training on massive datasets, LLMs demonstrate exceptional capabilities in semantic comprehension, inductive reasoning, and knowledge generalization \cite{liu2023pre}. Models like OpenAI's ChatGPT and Meta's Llama have overcome the limitations of traditional language models in terms of generation quality, logical coherence, and domain adaptability, thanks to their innovative architectures and optimized training \cite{touvron2023llama}. These advancements have not only driven transformative changes in natural language processing but also served as a cornerstone for the development of artificial general intelligence (AGI), heralding vast prospects for the application of intelligent technologies \cite{zhao2023brain}.

With the development of electronic systems and artificial intelligence, autonomous driving has emerged as a pivotal component of intelligent transportation systems, exhibiting substantial progression \cite{kiran2021deep}. LLMs, with their powerful language comprehension capabilities, provide innovative pathways to address the key challenges confronted by autonomous driving systems \cite{wu2024prospective} \cite{cui2024large} \cite{cui2024survey}. Particularly in dynamic and complex traffic scenarios, LLMs demonstrate significant improvements in tackling limitations such as insufficient environmental understanding and poor algorithmic interpretability in existing systems \cite{zhao2024survey} \cite{chen2024end}.

While LLMs excel in real-time responsiveness, their limitations in deep reasoning constrain performance on complex tasks \cite{huang2022towards}. Chain-of-Thought (CoT) addresses this by simulating human cognition through sequential reasoning steps, significantly enhancing model capabilities \cite{wei2022chain} \cite{xiang2025towards}. Widely recognized in both academic and industrial circles, recent reasoning large models leveraging CoT, such as OpenAI o1 and DeepSeek-R1, have demonstrated expert-level performance in mathematics and programming, substantially outperforming other LLMs \cite{jaech2024openai} \cite{guo2025deepseek}. Researchers are increasingly exploring CoT applications in autonomous driving, embodied AI, medicine, and finance \cite{manas2024cot} \cite{wang2024drivecot} \cite{liu2024medcot} \cite{deng2024leveraging}.

\begin{figure*}[htbp]
\centering
\includegraphics[width=6.4in]{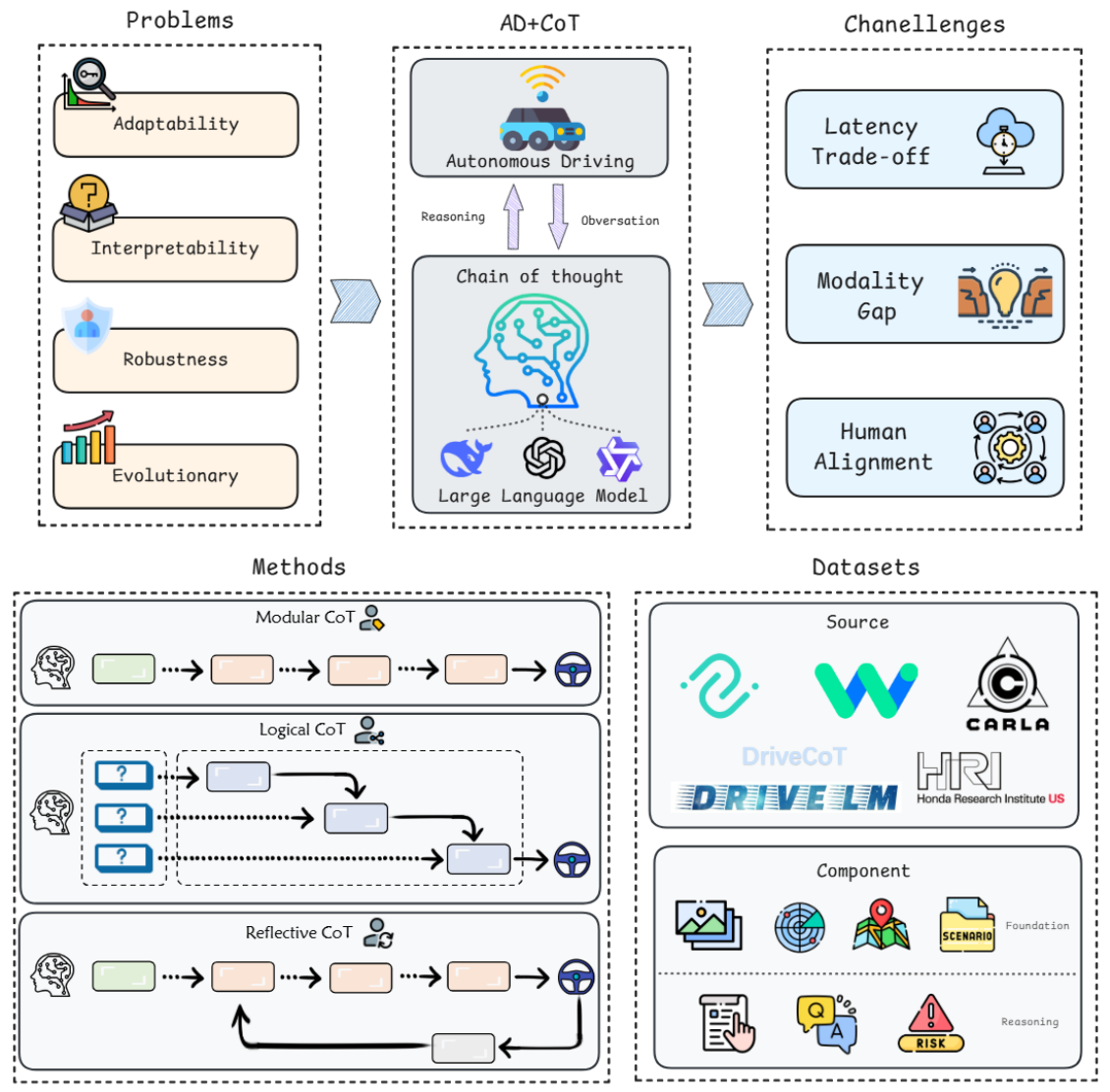}
\caption{It presents an overview of Chain-of-Thought for autonomous driving, outlining the motivations, framework, methodological paradigms, key challenges, and supporting datasets involved in this emerging research direction.}
\label{fig1} 
\end{figure*}

In the realm of autonomous driving, Chain-of-Thought technology, owing to its high alignment with human driving cognition, holds great potential to drive innovative breakthroughs in autonomous systems \cite{wang2024drivecot} \cite{chen2024m} . By simulating human drivers' decision-making processes through structured reasoning mechanisms, this technology significantly enhances a system's capability to handle complex scenarios \cite{fu2024drive} \cite{ma2024learning} \cite{tian2024drivevlm}. At its core, CoT decomposes driving tasks into interpretable, multi-step reasoning processes, enabling the system to make rational decisions through logical reasoning, just like an experienced human driver \cite{nie2024reason2drive}.

Although some review studies have explored the applications of LLMs, vision-language models (VLMs), and multimodal large models (MLMs) in autonomous driving, with some research touching upon chain-of-thought techniques \cite{cui2024survey} \cite{yang2023llm4drive}  \cite{gao2024survey} \cite{zhou2024vision}. To the best of our knowledge, none have focused specifically on how CoT technology advances autonomous driving. Therefore, there is a notable absence of a comprehensive review that consolidates and analyzes the application of CoT methods in autonomous driving. Most current reviews on CoT primarily analyze its underlying technical principles within large models \cite{chen2025towards} \cite{yu2023towards}. In contrast, this work provides a structured synthesis of the cutting-edge advancements of CoT technology in the field of autonomous driving. Beyond methodical categorization and dataset analysis, it critically delves into existing technical bottlenecks and proposes forward-looking research directions. Through this systematic  perspective, the aim is to provide readers with pioneering insights into how CoT is transforming autonomous driving.

The main contributions of this work can be summarized as follows:
\begin{enumerate}
    \item We present a comprehensive review of chain-of-thought reasoning in autonomous driving and categorize existing research along two dimensions.
    
    \item We conduct an exhaustive evaluation of existing autonomous driving reasoning datasets used for CoT, providing a comparative analysis of their characteristics and applicability.
    
    \item We identify and analyze promising applications of CoT methods in autonomous driving, while proposing concrete directions for future research.

    \item We provide an in-depth discussion of the core advantages, critical challenges, and research gaps in this field.

\end{enumerate}

The structure of this paper is as follows: Section II presents the background and foundational concepts of the study; Section III categorizes the Chain-of-Thought methods applied to autonomous driving; Section IV introduces the relevant datasets for driving-related CoT; Section V discusses the application methods and current trends; and Section VI summarizes the future challenges and directions for the application of driving CoT.

\section{Foundations of Autonomous Driving CoT}
This section delves into the relevant background, exploring the foundational concepts in the following areas: autonomous driving (II-A), the application of large language models in autonomous driving (II-B), and Chain-of-Thought (II-C).
\subsection{Development of Autonomous Driving}
The evolution of autonomous driving is primarily manifested in the continuous innovation of system architecture \cite{o2018scalable}. In the early stages of development, modular autonomous driving predominated, decomposing  critical modules such as perception, prediction, decision-making, planning, and control into discrete modules that are connected sequentially \cite{hu2023planning}. This architecture demonstrated considerable advantages during development and debugging. However, inherent limitations exist in terms of the inter-module information transfer accuracy and collaborative optimization efficiency. With the significant progress in deep learning, end-to-end methods gradually became the focal point of research \cite{chen2024end}. The fundamental principle of this paradigm is to directly extract features from raw sensor data and generate vehicle control actions or trajectories based on these features \cite{zhang2021end}. It not only substantially simplifies the complexity of the system architecture but also improves overall performance \cite{pan2020imitation}. Despite its promise, the end-to-end paradigm still faces many challenges, such as limited interpretability and difficulties in handling long-tail scenarios \cite{cui2025sustainable}. Addressing these issues constitutes a fundamental requirement for the next generation of autonomous driving.

From the perspective of technological paradigm evolution, autonomous driving has advanced through three distinct stages: rule-driven, data-driven, and knowledge-driven \cite{li2023towards}. 

In the rule-driven paradigm, which relies on manually created logical rules to operate \cite{treiber2000congested}. While it offers strong interpretability, it demonstrates significant limitations when handling complex and dynamic traffic scenarios \cite{xiao2021rule}. Subsequently, data-driven methods used large-scale driving datasets and deep neural networks to simulate human driving behavior, adopting end-to-end large models through imitation learning to improve system adaptability \cite{shan2020reinforcement} \cite{wang2022high}. However, this paradigm introduces challenges such as data dependency, poor generalization, and reduced interpretability.

In recent years, knowledge-driven autonomous driving has gradually become a promising direction, effectively combining the advantages of rule-driven and data-driven approaches and incorporating insights into human cognitive processes \cite{wen2023dilu} \cite{zhang2024wisead}. Knowledge, in this context, represents a highly abstract and systematic representation of human understanding regarding driving scenarios and decision-making, encapsulating distilled experience and reasoning \cite{li2023towards}. The knowledge-driven paradigm first constructs an abstract “knowledge representation space” to encode high-level driving concepts, causal relationships, and traffic rules \cite{brachman2004knowledge} \cite{wang2022towards}. This paradigm integrates cutting-edge technologies such as LLMs, world models, and self-learning methods \cite{wen2023dilu} \cite{xing2024comprehensive} \cite{tu2025role}. By combining these elements, it equips autonomous systems with advanced reasoning and continual learning capabilities. Specific cognitive techniques, including chain of thought, reflection mechanisms, and curiosity exploration, play a key role in driving this knowledge-enhanced autonomy \cite{pathak2017curiosity} \cite{zhang2025enhancing}. The knowledge-driven paradigm establishes a new theoretical foundation and methodological framework for addressing issues like corner cases, while paving the way for the development of higher-level autonomous systems \cite{mei2024continuously}.

\begin{figure*}[htbp]
\centering
\includegraphics[width=6.4in]{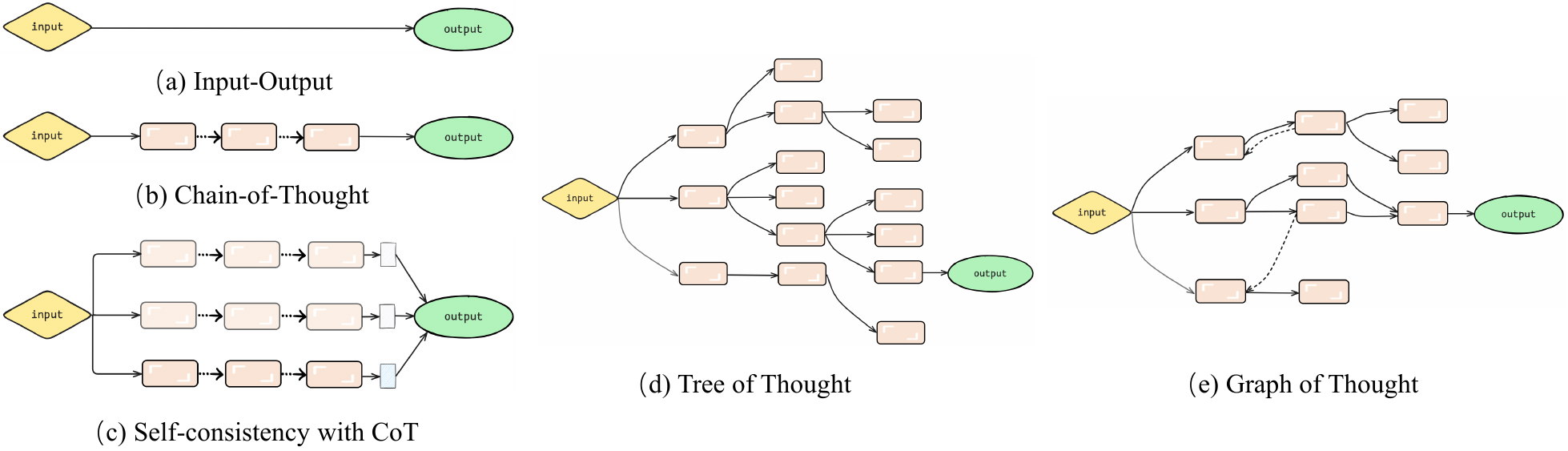}
\caption{It compares different variants of thought reasoning paradigms, from basic input-output to advanced forms like chain, tree, and graph of thought, highlighting the evolution of thought processes for complex problem-solving.}
\label{fig2} 
\end{figure*}

\subsection{LLMs for Autonomous Driving}

Transformer-based LLMs have ushered in a revolutionary era in artificial intelligence \cite{vaswani2017attention}. From early models like GPT-3 to current models such as GPT-4o, LLaMA 4, Deepseek-v3, and Gemini 2.5, LLMs exhibit unparalleled prowess in language understanding, knowledge acquisition, and logical reasoning \cite{touvron2023llama} \cite{team2023gemini} \cite{hurst2024gpt}  \cite{liu2024deepseek}. These advancements have profoundly impacted natural language processing (NLP) and computer vision (CV), and have opened up new transformative opportunities for autonomous driving technologies \cite{deng2024leveraging} \cite{li2024enhancing}.

The initial exploratory research on applying LLMs to autonomous driving primarily focused on converting sensor data and traffic rules into linguistic representations, leveraging LLMs for decision generation  \cite{fu2024drive} \cite{wen2023dilu}. It showcased LLMs' potential for static scene understanding and basic decision-making. However, when confronted with complex dynamic environments, the models still face challenges in generalization capability and real-time performance \cite{gao2024survey}.

The emergence of multimodal large language models (MLLMs), particularly vision language models (VLMs) and vision language action models (VLAs), has enabled deep fusion of heterogeneous sensor data including cameras, LiDAR, and HD maps \cite{zhang2024mm} \cite{li2025visual} \cite{zhou2025opendrivevla}. This breakthrough significantly enhances the system's capability for comprehensive, multi-dimensional environmental perception and integrated reasoning. Concurrently, as MLLMs' integration with autonomous driving systems deepens, the end-to-end paradigm further improves autonomous driving systems' overall performance \cite{jiang2025alphadrive}. The next generation of reasoning LLMs will exhibit significantly improved intelligence, adaptability, and generalization performance \cite{qu2025survey}.

\subsection{Theoretical Foundations of CoT}

LLMs have demonstrated remarkable emergent capabilities in contextual learning and complex reasoning tasks. Among these, the Chain-of-Thought reasoning technique has garnered significant attention due to its human-like cognitive characteristics \cite{wei2022chain}  \cite{dasgupta2022language}. It enables large models to solve novel problems through step-by-step reasoning processes, demonstrating capabilities that approximate human cognitive patterns \cite{wang2022towards}.

From a methodological perspective, Chain-of-Thought represents a structured multi-step reasoning paradigm that differs fundamentally from traditional intuitive Input-Output response mechanisms \cite{sprague2024cot}. It marks a cognitive advancement in AI interaction, shifting from mere result output to comprehension of the process. By establishing sequential reasoning processes, CoT allows models to perform multi-step derivations when addressing complex problems, ultimately generating well-justified explanations or decisions \cite{lightman2023let}. One distinctive feature of this approach is its recursive nature, where each reasoning step depends on the outcome of the preceding step, thereby forming a coherent logical chain.

While a standardized formulation of CoT reasoning has yet to be established, this study adopts a formalism similar to state transition to characterize the reasoning process. The recursive decomposition of each reasoning step in the thought chain is represented as the process of thought transition. Taking the most classic chain of thought as an example, the reasoning process is represented as a chain of thought transitions, formally described as follows:
\begin{align}
C := (P \xrightarrow{T_1} S_1) \odot (S_1 \xrightarrow{T_2} S_2) \odot \cdots \odot (S_{n-1} \xrightarrow{T_n} R)
\end{align}

\begin{itemize}
    \item $P$ is the initial task. $R$ is the final result.
    \item $T_1$, $T_2$, \dots, $T_n$ are the reasoning steps, i.e., the reasoning processes in the chain.
    \item $S_1$, $S_2$, \dots, $S_n$ are the intermediate states obtained after each reasoning step.
    \item $\odot$ denotes the symbol between reasoning tasks, indicating that each reasoning step depends on the result of the previous one.
\end{itemize}

The development of Chain-of-Thought technology has undergone several key stages. Initially proposed by DeepMind, the CoT guides LLMs to progressively demonstrate their reasoning processes, forming essential intermediate concepts \cite{wei2022chain}. The original research employed few-shot CoT, which provides manually designed examples containing both problems and their corresponding reasoning chains in the prompts to instruct models in generating similar reasoning processes \cite{brown2020language}. Subsequent research found that even simple prompts like "Let's think step by step" could effectively stimulate models' reasoning capabilities, leading to the emergence of Zero-shot CoT \cite{kojima2022large}. This approach significantly reduces the manual effort required for designing reasoning chains while improving scalability \cite{shaikh2022second}. Recognizing potential logical errors in reasoning chains generated solely through Zero-shot CoT, researchers proposed Auto-CoT, which automatically constructs reasoning demonstrations for large models through clustering and example sampling. Supervised CoT provides models with explicit reasoning steps to enhance their structured reasoning capabilities through supervision \cite{zhang2024supervised}.

Recent extensions to CoT methods have yielded several innovative variants optimized for different aspects of complex reasoning. These include Self-consistency with CoT, which improves robustness by integrating multiple reasoning paths \cite{wang2022self}; Tree of Thought (ToT), which supports branching exploration \cite{yao2023tree}; and the Graph of Thought (GoT) framework that enables graph-structured reasoning \cite{besta2024graph}. Each advancement contributes to expanding the application scope and effectiveness of reasoning-capable large language models. Notably, Visual CoT employs textual representations of visual content as an intermediary between visual inputs and reasoning processes \cite{shao2024visual}. CoT applications in visual domains show particular promise for multimodal reasoning tasks, such as critical challenges in autonomous driving scenarios that require integrating visual perception with high-level reasoning \cite{zhang2023multimodal}.

This section aims to provide a concise background introduction to Chain-of-Thought and summarize its subsequent developments, offering readers a clear framework for understanding these key concepts and their applications in autonomous driving.


\begin{table*}[!p]
\setlength{\tabcolsep}{4pt}
\hspace*{-0.3cm}
\centering
\caption{Comparative Analysis of Autonomous Driving CoT Models}

{\footnotesize 
\textbf{Abbreviations:} SF / MF = Single-frame / Multi-frame, SV / MV = Single-view / Multi-view, 2DKS = 2D position / Kinematic state, \par
TI = Task instruction, EH = Ego history, Mem = Memory}
\vspace{0.2cm}
\label{table_cognitive_models}
\renewcommand{\arraystretch}{1.6}
\begin{tabular}{lllll}
\toprule
\textbf{Name} & \textbf{Year} & \textbf{Task} & \textbf{Pipeline} & \textbf{Chain-of-thought cognitive process } \\
\midrule
Agent-Driver\cite{mao2023language} & 2023 & Perception & Reflective & SF-MV → Key object detection → High-level intent → Trajectory $\circlearrowleft$ Refine, Mem \\
Dolphins\cite{ma2024dolphins} & 2024 & Perception & Logical & MF-SV, Question → GCoT-fine-tuning → Perception answer  \\
RIV-CoT\cite{corbière2025retrievalbasedinterleavedvisualchainofthought}
 & 2025 & Perception & Logical & SF-SV, Question → Bounding box → Image crop → Perception answer\\
DriveAgent\cite{hou2025driveagent} & 2025 & Perception & Modular & MF-SV, LiDAR → Description → Vehicle reasoning → Scene analysis → Perception \\
AgentThink\cite{qian2025agentthink} & 2025 & Perception & Logical & SF-MV, Question → Tool use → Uncertainty flag → Perception answer\\
Reason2Drive\cite{nie2024reason2drive} & 2024 & Prediction & Modular & MF-SV, Question → Perception answer → Prediction answer → Motion visualization \\
Motion-LLaVA\cite{li2024womd} & 2024 & Prediction & Modular & 2DKS, Question → Aggregated in-context reasoning → Prediction answer \\
LC-LLM\cite{peng2025lc} & 2025 & Prediction & Logical & 2DKS, EH → Feature detection → Intention → Motion prediction, Explanation \\
CoT-Drive\cite{liao2025cot} & 2025 & Prediction & Logical & 2DKS → Background → Interaction analysis → Risk assessment → Motion prediction \\
SenseRAG\cite{luo2025senserag} & 2025 & Prediction & Logical & SF-MV, LiDAR, Structured data → Data injection → RAG → Motion Prediction \\
GPT-Driver\cite{mao2023gpt} & 2023 & Planning & Modular & 2DKS, EH, TI → Key object detection → Interaction prediction → Trajectory  \\
PlanAgent\cite{zheng2024planagent} & 2024 & Planning & Reflective & MF-MV, 2DKS → Global, local info → Scene description → Planning code $\circlearrowleft$ Refine  \\
DriveVLM\cite{tian2024drivevlm} & 2024 & Planning & Logical & MF-MV → Scene description → Scene Analysis → Hierarchical Planning → Trajectory  \\
RDA-Driver\cite{huang2024making} & 2024 & Planning & Modular & SF-MV, EH → Key object detection and prediction → High-level intent → Trajectory   \\
AlphaDrive\cite{jiang2025alphadrive} & 2025 & Planning & Modular & MF-SV → Key object detection → High-level intent \\
CALMM-Drive\cite{yao2024calmm} & 2025 & Planning & Logical & BEV, EH, TI → Top-K action, Confidence → Trajectory → Hierarchical refinement\\
LanguageMPC\cite{sha2310languagempc} & 2023 & Decision & Logical & 2DKS, EH, TI → Vehicle detection → Situational awareness → Meta action  \\
DiLu\cite{wen2023dilu} & 2023 & Decision & Reflective & 2DKS → Scene description → Meta action $\circlearrowleft$ Refine, Mem \\
DriveMLM\cite{wang2023drivemlm} & 2023 & Decision & Logical & MF-MV, TI, LiDAR → Linguistic description → Speed-path decision  \\
Receive-Reason-React\cite{cui2024receive} & 2024 & Decision & Reflective & 2DKS, TI, In-cabin info → Scene description → Explanation → Meta action $\circlearrowleft$ Mem\\
SafeDrive\cite{zhou2024safedrive} & 2024 & Decision & Reflective & 2DKS, TI → Risk evaluation → Key object detection → Meta action $\circlearrowleft$ Refine, Mem \\
KOMA\cite{jiang2024koma} & 2024 & Decision & Reflective & 2DKS → Scene description → Goal → Planning → Meta action $\circlearrowleft$ Refine, Mem \\
LeapAD\cite{mei2024continuously} & 2024 & Decision & Reflective & SF-MV → Scene description → Dual-process $\circlearrowleft$ Refine, Mem  \\
LeapVAD\cite{ma2025leapvad} & 2025 & Decision & Reflective & MF-MV → Scene description, ACC-ACT similarity → Dual-process $\circlearrowleft$ Refine, Mem  \\
CoDrivingLLM\cite{fang2025towards} & 2025 & Decision & Reflective & 2DKS → State perception → Intent sharing → Negotiation → Meta action $\circlearrowleft$ Mem \\
Actor-Reasoner\cite{fang2025interact} & 2025 & Decision & Reflective & 2DKS, TI, Mem database → Intent prediction → Driving style → Meta action $\circlearrowleft$ Mem\\
CoT-VLM4Tar\cite{ren2025cot} & 2025 & Decision & Logical & SF-SV → Situation classification → Scene Analysis → High-level intent → Meta action \\
LLM-Driver\cite{chen2024driving} & 2024 & E2E & Modular & 2DKS → Scene vectors grounding → Prediction answer, Control \\
PKRD-CoT\cite{luo2024pkrd} & 2024 & E2E & Reflective & SF-MV → Scene description → Object detection → High-level intent $\circlearrowleft$ Mem\\
LMDrive\cite{shao2024lmdrive} & 2024 & E2E & Modular & MF-MV, TI, LiDAR → Feature detection → Trajectory → PID control \\
WiseAD\cite{zhang2024wisead} & 2024 & E2E & Modular & MF-SV, TI, Question → Answer (Scene description, Risk analysis...) + Trajectory \\
DriveCoT\cite{wang2024drivecot} & 2024 & E2E & Logical & MF-MV → Multidimensional prediction → Logical decision → Meta action \\
Senna\cite{jiang2024senna} & 2024 & E2E & Logical & SF-MV, TI → Meta-action → Perception → Motion prediction → Trajectory \\
EMMA\cite{hwang2024emma} & 2024 & E2E & Modular & SF-MV, EH, TI → Scene description → Key object description → Meta action \\
OpenEMMA\cite{xing2025openemma} & 2025 & E2E & Modular & SF-SV, EH → Key object, Scene description, High-level intent → Trajectory\\
LightEMMA\cite{qiao2025lightemma} & 2025 & E2E & Modular & SF-SV, EH → Scene description → High-level intent → Trajectory\\
ORION\cite{fu2025orion} & 2025 & E2E & Reflective & SF-MV, TI → Feature extraction → Scene analysis → Meta action → Trajectory $\circlearrowleft$ Mem\\
DriveLM\cite{sima2024drivelm} & 2025 & E2E & Modular & SF-SV, Question → Perception → Prediction → High-level intent → Trajectory \\
See2DriveX\cite{zhao2025sce2drivex} & 2025 & E2E & Modular & MF-MV, 2DKS, TI, BEV → Scene description → Meta action → Trajectory → Control\\
PRIMEDrive-CoT\cite{mandalika2025primedrive} & 2025 & E2E & Logical & SF-MV, LiDAR → Uncertainty/Risk → Interaction → Logical decision → Meta action\\
LangCoop\cite{gao2025langcoop} & 2025 & E2E & Logical & SF-SV → Scene description → High-level intent → LangPack integration → Control\\
X-Driver\cite{liu2025x} & 2025 & E2E & Logical & SF-SV, TI → Object detection, Traffic sign, Lane info → Trajectory \\

\bottomrule
\end{tabular}
\end{table*}

\section{CoT Methods for Autonomous Driving}
The Chain-of-Thought method is gradually being applied in autonomous driving to enhance the system's reasoning ability and decision-making level in complex environments. This section systematically classifies existing methods from two dimensions: structural characteristics(III-A) and task domain(III-B). The task domain perspective focuses on the specific applications of CoT, while the structural characteristics perspective reveals the differences in CoT's structural design. This dual classification framework helps to comprehensively review the current research outcomes.

\subsection{Pipeline Paradigms}
This paper will conduct a thorough analysis of autonomous driving CoT's structural characteristics from the pipeline paradigm perspective.

\paragraph{\textbf{Modular Driving CoT}}

The Modular Driving CoT pipeline adheres to the hierarchical architecture of autonomous driving, decomposing the driving task into multiple independent submodules like perception, decision-making, planning, and control. Each submodule is responsible for specific reasoning tasks and sequentially transmits its outputs downstream, thereby achieving driving functionality in complex scenarios. Through hierarchical propagation, raw sensor data is progressively transformed into final driving commands. This modular design not only enhances the system's interpretability but also enables independent optimization of individual modules, allowing flexible adaptation to diverse scenario requirements.
The mathematical formulation of the Modular Driving CoT pipeline is as follows:
\begin{align}
C_m := \bigodot_{i=1}^n \bigl(S_{i-1} \xrightarrow{T_i} S_i\bigr)\;\to\;R
\end{align}

\begin{figure}[htbp]
\centering
\includegraphics[width=3.2in]{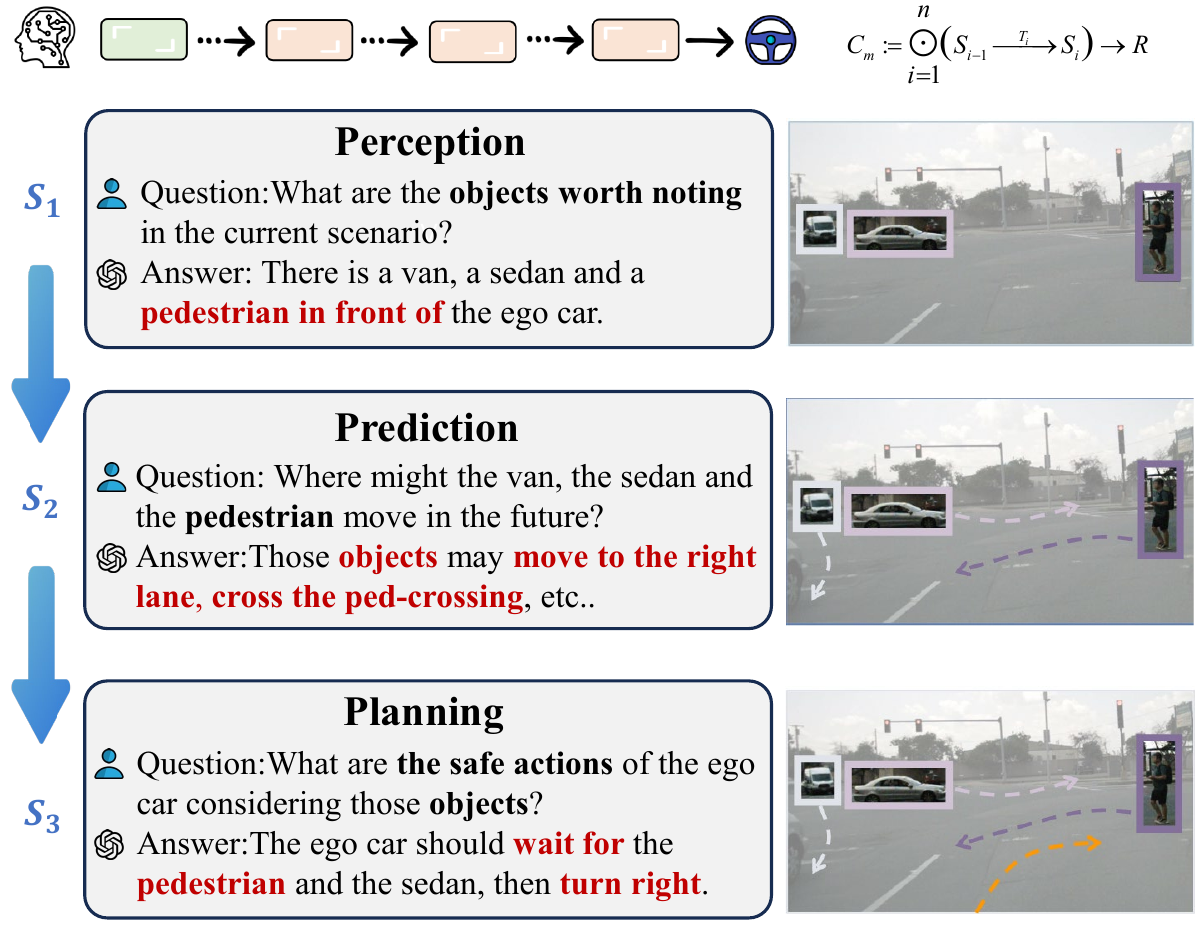}
\caption{As a representative of Modular Driving CoT, DriveLM decomposes driving tasks into subtasks including perception, prediction, and planning \cite{sima2024drivelm}. CoT provides explicit reasoning processes for decision-making.}
\label{fig3} 
\end{figure}

The first module takes perceptual information as input, where $T_i$ denotes the thought transition process of the i-th submodule,$S_i$ represents the intermediate state of the i-th module. Each module processes the input state $S_{i-1}$ via $T_i$ and outputs $S_i$, ultimately generating the driving output $R$.

The pipeline adopts a modular design, enabling each module of the thought chain to independently process relevant data while coordinating through a unified system. This modular pipeline effectively reduces system coupling, enhancing operational transparency and traceability while facilitating development and debugging. However, as each module pursues distinct optimization objectives, joint training struggles to achieve global optimization. Additionally, multi-stage information transmission tends to accumulate latency and errors.
\paragraph{\textbf{Logical Driving CoT}}

\begin{figure*}[htbp]
\centering
\includegraphics[width=6.4in]{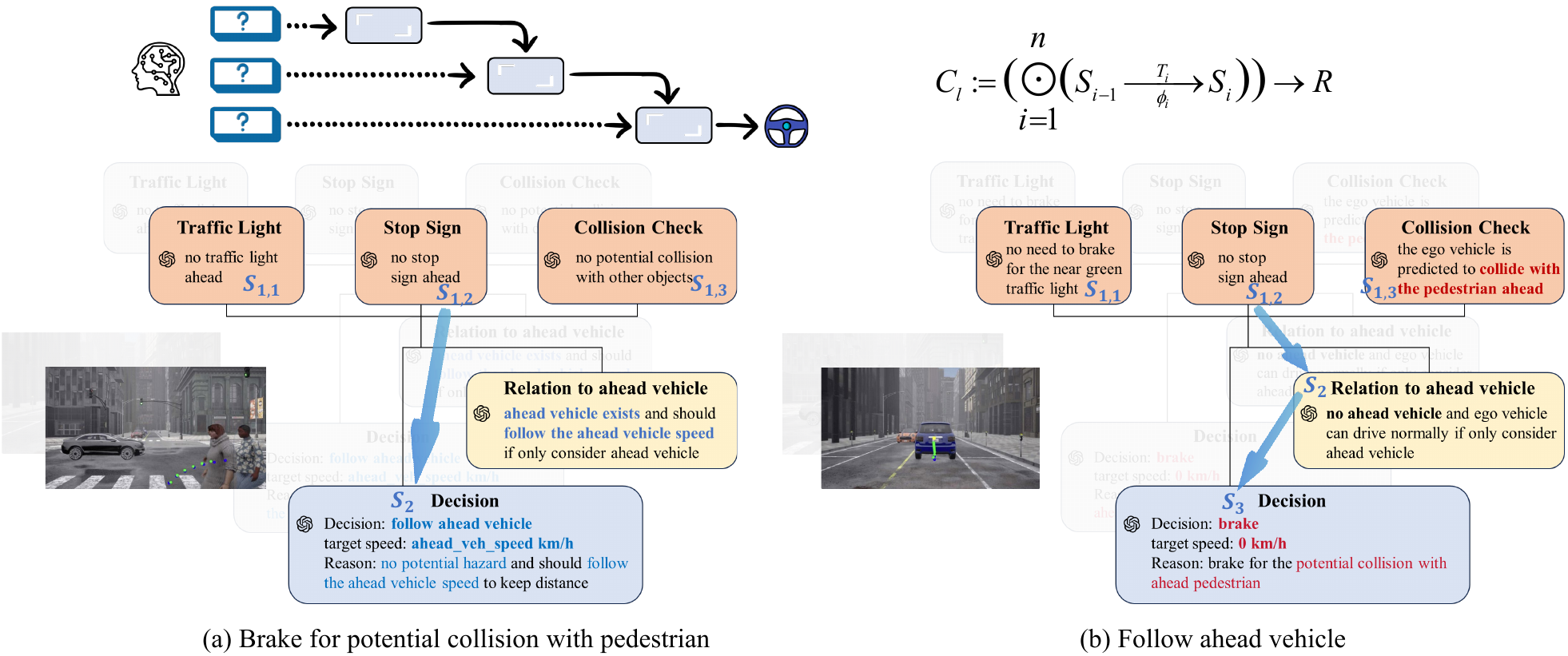}
\caption{PRIMEDrive-CoT evaluates collision probability, occlusion conditions, and unknown object risks through a hierarchical triple-risk verification process, constructing a reasoning framework based on Logical Driving CoT \cite{mandalika2025primedrive}.}
\label{fig4} 
\end{figure*}
Unlike Modular Driving CoT which strictly follows autonomous driving's hierarchical architecture, Logical Driving CoT adopts a fundamentally different approach during the Problem Decomposition phase by breaking down driving tasks into multiple logically interconnected sub-problems rather than modular components \cite{wang2024drivecot} \cite{ren2025cot}. In the subsequent Subproblem Solving phase, the Logical Driving CoT methodology places greater emphasis on constraining the reasoning process through rigorous logical judgments. It introduces a more sophisticated logical constraint mechanism where thought processes and logical chains collaboratively govern the system's operations, ultimately generating driving decisions through a series of logical steps. The pipeline constructs multi-branch reasoning chains that perform cognitive reasoning, conditional judgments, or cost evaluations at each logical node to optimize candidate strategies. The Logical Driving CoT can be formally expressed as follows:
\begin{align}
C_l := 
\Bigl(\bigodot_{i=1}^n \bigl(S_{i-1}\xrightarrow[\phi_i]{T_i}S_i\bigr)\Bigr)
\;\to\;R
\end{align}

$\phi_i$ denotes a logical operator that constrains the reasoning process. The solution to each sub-problem must simultaneously satisfy both the thought transition information $T_i$ and the logical constraint $\phi_i$.

The pipeline also incorporates subtasks like collision-risk estimation and stop sign checks, utilizing Least-to-Most Prompting \cite{zhou2022least}. This strategy guides the model through progressively complex questions, reflecting the step-by-step reasoning process humans use. Taking DriveCoT as a case in point, the system first detects obstacles ahead and analyzes their motion states and relative positions using real-time perception data. Through this logical reasoning chain, it then determines appropriate strategies like deceleration, stopping, or evasive maneuvers. This logic-driven reasoning approach significantly enhances decision-making accuracy and stability \cite{mandalika2025primedrive}. Particularly in high-risk scenarios, it effectively reduces the likelihood of misjudgments and decision latency.

\paragraph{\textbf{Reflective Driving CoT}}
The Reflective Driving CoT builds upon the previous two types of CoT reasoning pipelines by introducing the reflective update mechanism. By evaluating the discrepancy between execution results and pre-stored experiences, it triggers self-reflection at the chain's terminal, enabling system self-iteration, error correction, and memory bank updates.

This pipeline transforms the unidirectional thought chain into a bidirectional cognitive loop, equipping the system with causal reasoning and knowledge transfer capabilities. Consequently, in out-of-distribution (OOD) scenarios, the system can generate adaptive strategies through logical inference, effectively handling complex and dynamic driving environments. The formulation is as follows:
\begin{align}
C_r := 
\Bigl(\bigodot_{i=1}^n \bigl(S_{i-1}\xrightarrow{T_i}S_i\bigr)\Bigr)
\;\odot\;
\Bigl(S_n \circlearrowleft  S_0'\Bigr)
\;\to\;R
\end{align}

Among them, $\circlearrowleft$ represents the reflective feedback process. $S'$ is the reflective information generated by the trajectory function $\tau$, which includes an evaluation of historical states and thought transition processes.

The pipeline of Reflective Driving CoT possesses continual learning and long-term optimization capabilities, as exemplified by the typical system Dilu \cite{wen2023dilu}. By analyzing recorded decision sequences, the reflection module identifies potentially unsafe or inaccurate decisions. Subsequently, LLM is employed to correct these erroneous decisions, and the revised reasoning process along with the correct decision outcomes is stored in the memory module. The memory module utilizes a vector database to store past driving scenario experiences, including decision prompts, reasoning processes, and other metadata, in vectorized form for future retrieval. Corrected driving decision information is also updated in the memory module. This pipeline achieves human-like reflective reasoning, enabling the system to continuously self-learn and evolve, thereby progressively improving driving performance.
\begin{figure*}[htbp]
\centering
\includegraphics[width=6.4in]{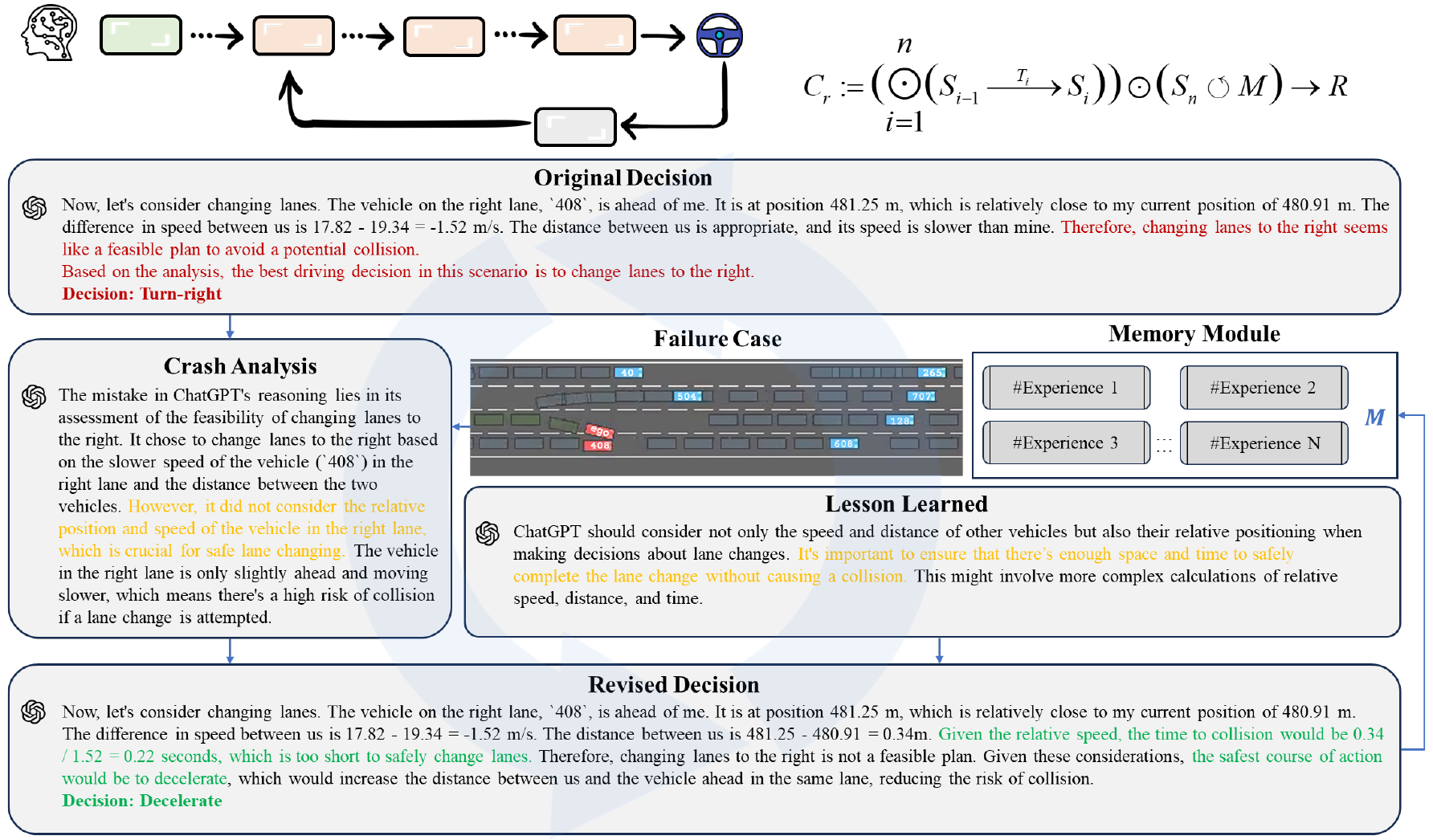}
\caption{Dilu employs Reflective Driving CoT, integrating memory recall and introspection into the reasoning process \cite{wen2023dilu}. The reflection module combines real-time cognitive data from the current scenario with selective past experiences retrieved from the memory bank, jointly feeding them into the LLMs for response decoding and action execution.}
\label{fig5} 
\end{figure*}

\subsection{Task-Oriented Domains}
In the diverse applications of autonomous driving, the CoT methods can be divided into four main areas based on the different task domains: perception and understanding, prediction and planning, decision-making and control, and end-to-end systems. The previous investigation of CoT methods in autonomous driving systems is grounded in the pipeline paradigm. The following content will offer a more comprehensive investigation within the task domain framework.

\paragraph{\textbf{Perception and Understanding}}
Complex environmental interpretation forms the foundation of autonomous driving. CoT reasoning offers a novel perspective for multi-source data fusion and dynamic environment comprehension in perception systems. Dolphins proposed a Grounded Chain-of-Thought (GCoT)-based multimodal conversational driving assistant, trained on driving-specific instruction data \cite{ma2024dolphins}. It begins with summarizing the core content of the image, followed by accurately identifying the objects within the scene and describing their spatial relationships. The process concludes with a thorough analysis of the driving perception task. By integrating these stages, a standardized GCoT response format can be established, delivering a more flexible and intelligent perception solution. RIV-CoT employs a Retrieval-Based Interleaved Visual Chain-of-Thought method\cite{corbière2025retrievalbasedinterleavedvisualchainofthought}. Different from other methods that rely solely on text CoT, RIV-CoT embeds visual evidence directly into the multi-step reasoning chain through real-time image cropping, effectively mitigating hallucination phenomena in large models . DriveAgent employs a structured four-module sequential framework, comprising descriptive analysis, vehicle reasoning, environmental reasoning, and response generation, to address both environment-level and vehicle-level tasks \cite{hou2025driveagent}. The framework employs multi-agent collaboration to fuse LiDAR-camera data through cross-modal consistency validation, enabling robust scene understanding via structured reasoning and iterative refinement.

\paragraph{\textbf{Prediction and Planning}}
Trajectory prediction and planning are central to autonomous driving, using sensory and historical data to forecast and generate motion. CoT reasoning offers a novel framework for hierarchical inference, markedly improving performance.

In the domain of prediction using CoT methods, LC-LLM stands out as the first to frame trajectory forecasting as a language modeling problem \cite{peng2025lc}. It innovatively integrates three types of domain knowledge: traffic rules, traditional lane-changing models, and driving scene characteristics. Leveraging CoT reasoning, it establishes a dual-layer reasoning framework where feature annotation and latent behavior prediction operate concurrently, enhancing both prediction accuracy and interpretability. CoT-Drive adopts a teacher-student architecture, where the teacher LLM's reasoning process is optimized via chain-of-thought and then distilled into a lightweight student language model, achieving a balance between coherent reasoning and edge-device deployment \cite{liao2025cot}. Specifically, in the teacher model's CoT process, reasoning proceeds through four fundamental stages, beginning with background and statistical analysis, progressing through interactive behavior examination and risk evaluation, and culminating in prediction formulation.This thought-guided approach produces semantic descriptions that comply with traffic rules and human driving logic, ensuring precision and explainability. Unlike methods reliant on predefined tool libraries or template queries, SenseRAG combines CoT with multimodal Retrieval Augmented Generation(RAG), dynamically acquiring external knowledge to improve prediction performance \cite{luo2025senserag}. Meanwhile, Reason2Drive enhances LLMs' reasoning capabilities by integrating object-level perception, enabling predictive analysis of accident-prone scenarios or hazardous situations \cite{nie2024reason2drive}.

In the planning section, DriveVLM, building upon Qwen-VL and incorporating CoT reasoning, proposes a progressive planning framework of "scene description – scene analysis – hierarchical planning"\cite{tian2024drivevlm} \cite{bai2023qwen}. This framework initially evaluates environmental factors during the scene description phase and obtains target objects of various categories along with their bounding box information through key object detection. Subsequently, in the scene analysis phase, it extracts static attributes, motion states, and special behaviors of the objects. The system then transmits the scene summary generated from prior reasoning to the hierarchical planning module. During the hierarchical planning phase, leveraging the preceding reasoning results, the system generates meta-actions, decision descriptions, and trajectory nodes based on prompt information, optimizing the trajectory to achieve precise path planning. GPT-Driver enhances path planning accuracy and efficiency by linguistically formulating the motion planning problem through thought annotations of future collision probabilities \cite{mao2023gpt}. RDA-Driver introduces a reasoning-decision alignment mechanism, designing consistency constraint loss and contrastive function loss to help the model comprehend the involved logical reasoning, ensuring logical consistency between the model's explanations and conclusions while correcting inconsistencies between CoT reasoning information and planning results \cite{huang2024making}. CoT establishes causal relationships to predict and analyze potential event sequences, thereby adjusting behaviors accordingly to effectively plan driving routes and actions. PlanAgent employs hierarchical multi-round CoT reasoning to generate planner code for the Intelligent Driver Model (IDM), reasoning about the current scenario across three levels: scene understanding, motion commands, and code generation \cite{zheng2024planagent} \cite{jiang2024survey}. CALMM-Drive flexibly modifies strategies based on scene classification to achieve multi-step reasoning, deriving Top-K decisions with attached confidence levels \cite{yao2024calmm}. The multi-decision output mechanism effectively reduces uncertainty in the CoT reasoning process while significantly mitigating risks posed by single erroneous decisions through the integration of multiple logical strategies.

\paragraph{\textbf{Decision-Making and Control}}

Decision-making and control represent another core challenge in autonomous driving, responsible for translating high-level intentions into concrete operations. CoT enables autonomous vehicles to generate optimal control commands and offer human-interpretable decision justifications, aligning with logical human reasoning processes.

LanguageMPC utilizes three predefined tool libraries: scenario encoding, action guidance, and confidence adjustment \cite{sha2310languagempc}. It transforms linguistic decisions from LLMs into structured representations compatible with model predictive control (MPC) requirements \cite{kouvaritakis2016model}. These include observation matrices, action bias vectors, and dynamic weight matrices, enabling adaptive attention allocation and contextual understanding. DRIVEMLM combines the chain of thought with a traffic knowledge base, emphasizing consistency in decision-making and planning \cite{wang2023drivemlm}. Receive-Reason-React employs the chain of thought as a guiding signal to ensure alignment with human-like reasoning \cite{cui2024receive}. It integrates in-cabin monitoring, driver verbal commands, and contextual reasoning to enhance safety and robustness while delivering a personalized driving experience. CoT-VLM4Tar proposes a four-stage chain of thought for traffic anomaly detection, guiding the VLM through a precise reasoning process—from phenomenon-level scene classification to analytical reasoning, high-dimensional intervention measures, and finally, structured formatting of key vehicle movements translating VLM-generated solutions into executable commands \cite{ren2025cot}. SafeDrive combines natural language scene descriptions, a dynamic Driver Risk Field (DRF), and historical memory to drive LLM-based three-stage reasoning \cite{zhou2024safedrive}. It forces the LLM to generate actions within the hard safety thresholds of the risk module, with each decision accompanied by a risk traceability report to ensure system safety.

Specifically, the dual-decision process is one of the unique reasoning systems within the thinking architecture \cite{evans1984heuristic} \cite{li2025system}. LeapAD constructs a parallel fast-slow dual-decision process, simulating the coexistence of rationality and intuition in the human brain \cite{mei2024continuously}. The slow Analytic Process (System-II) mimics human rationality to generate reliable driving decisions and enables post-incident reflection. Meanwhile, the fast Heuristic Process (System-I) employs a lightweight language model to retrieve reference knowledge from a memory bank, enabling real-time reasoning and supporting cloud-edge collaborative deployment. Building upon this, LeapVAD introduces multi-frame scene summarization and "steering-braking" contrastive learning to enhance similar sample retrieval and scene comprehension \cite{ma2025leapvad}. In cooperative driving automation deployments, the Actor-Reasoner framework similarly adopts a dual-track cognitive structure—combining retrieval-based fast reasoning and systematic slow reasoning \cite{fang2025interact}. The Reasoner performs procedural predictions based on scene descriptions and experiential summaries while refining driving style elements. The Actor integrates the Reasoner’s outputs with dual-layer memory retrieval to achieve rapid, lightweight reflexes. Driving style and expressive Human-Machine Interface (eHMI) information serve as core intermediate outputs of the CoT, enabling efficient and safe decision-making in multi-vehicle coordination.

\paragraph{\textbf{End-to-End Autonomous Driving}}
End-to-end autonomous driving simplifies system architecture through unified modeling from perception to control, but it lacks interpretability. Introducing CoT reasoning helps construct a unified cognitive representation and transparent reasoning path, thereby enhancing the system’s verifiability and explainability.

PKRD-CoT leverages Zero-shot CoT to enable MLLMs to perform autonomous driving tasks \cite{luo2024pkrd}. The system implements four fundamental capabilities processing architecture tailored to autonomous driving, consisting of perception, knowledge, reasoning, and decision-making. It allows for system deployment without requiring pretraining, while simultaneously ensuring interpretable decision processes. Similarly, Waymo's EMMA explicitly prompts models to perform step-by-step visual information reasoning, achieving synergy between explicit rationale generation and data-driven models \cite{hwang2024emma}. Reproducible and enhanced follow-up work, such as OpenEMMA and LightEMMA incorporates explicit CoT reasoning. This includes components like high-level intent command, driving scene understanding, and major object detection, which work together to accurately identify objects, plan trajectories, and support rational driving decisions \cite{xing2025openemma} \cite{qiao2025lightemma}. DriveCoT-Agent decomposes end-to-end driving tasks into sequential logical steps—such as collision prediction, traffic sign recognition, relation to ahead vehicle, and road recognition \cite{wang2024drivecot}. Within this reasoning chain, the system first evaluates prediction outputs from the network to determine whether potential hazards triggering emergency braking exist. If none are detected, it proceeds to analyze the relative dynamics between the ego vehicle and preceding vehicles, ultimately generating the final driving decisions. PRIMEDrive-CoT specifically integrates uncertainty and risk-aware reasoning for interactive object learning, generating human-aligned decision explanations through risk prediction and uncertainty quantification \cite{mandalika2025primedrive}.

Beyond these, DriveLM implements graph-structured reasoning via Graph Visual Question Answering (GVQA), organizing QA pairs into directed acyclic graphs \cite{sima2024drivelm}. Its decision-making pipeline follows a chained workflow, Perception → 
Prediction → Planning → Behavior → Motion, where each QA pair incorporates parent-node context to explicitly propagate reasoning-chain information, mimicking human cognitive processes. What's more, WiseAD and LMDrive adopt an implicit CoT design, which systematically integrates structured CoT reasoning traces into LLMs training \cite{zhang2024wisead} \cite{shao2024lmdrive}. ORION introduces a vision-language instructed action generation paradigm, utilizing QT-Former for scene modeling and employing a generative planner to map semantic information from LLMs into executable trajectories \cite{fu2025orion}. It bridges semantic and numerical spaces via generative models. Senna combines the commonsense reasoning of VLMs with motion intention prediction and meta-action planning \cite{jiang2024senna}. Leveraging meta-action features as high-level guidance, it facilitates accurate reasoning. See2DriveX reconstructs the inherent implicit cognitive chain of human driving \cite{zhao2025sce2drivex}. It enables step-by-step deduction from scene understanding, meta-action reasoning, behavior interpretation analysis, motion planning and control, narrowing the gap between autonomous driving and human thought processes. LangCoop uses language as an intermediate representation for collaborative driving, enabling step-by-step sharing of reasoning states across vehicles \cite{gao2025langcoop}. Each vehicle derives preliminary driving intentions via CoT reasoning. The LangPack module then packages and shares knowledge, ultimately feeding back to individual vehicles' motion planning.

\section{Datasets}
Open-source datasets serve as essential resources that fuel the innovation in autonomous driving, furnishing extensive data for system training, testing, and validation \cite{liu2021survey}. With technological evolution, dataset construction has progressively evolved beyond focusing solely on perception and behavior to integrate semantic cognition \cite{liu2024survey}. As a fundamental medium of human intelligence, linguistic elements are increasingly integrated to enhance models' reasoning capabilities \cite{li2019aads}. Training on large-scale datasets, enriched with semantic reasoning information, provides a transformative pathway for developing autonomous agents with human-like cognitive capabilities \cite{xie2025vlms}. Recent advancements demonstrate the growing implementation of explicit reasoning representations, particularly CoT mechanisms, which show promising potential in modeling complex decision-making processes \cite{liu2024survey}. This section provides a systematic review of foundational autonomous driving datasets, cognition-augmented datasets, and relevant evaluation metrics. It is followed by a comparative analysis of their characteristics, summarized in the table below.


\begin{table*}[!ht]
\centering
\caption{Autonomous Driving Datasets Incorporating Cognitive Intelligence}
\label{table_multimodal_datasets}
\renewcommand{\arraystretch}{1.5}
\begin{tabular}{lcccccc}
\toprule
\textbf{Name} & \textbf{Year} & \textbf{Source} & \textbf{Cognitive Data Type} & \textbf{Size} & \textbf{Tasks} \\
\midrule
\textbf{Talk2Car \cite{deruyttere2019talk2car}} & 2019 & nuScenes & VQA & 850 videos, 11,959 commands & Perception \\
\textbf{nuScenes-QA \cite{qian2024nuscenes}} & 2024 & nuScenes & VQA & 28K frames, 459,941 QAs & Perception \\
\textbf{Talk2BEV \cite{choudhary2024talk2bev}} & 2024 & nuScenes & VQA & 1K BEV scenarios, 20K QAs & Perception \\
\textbf{NuScenes-MQA \cite{inoue2024nuscenes}} & 2024 & nuScenes & Markup-QA & 34,149 scenarios, 1,459,933 QA pairs & Perception \\
\textbf{Reason2Drive \cite{nie2024reason2drive}} & 2024 & nuScenes, Waymo, ONCE & QA & 420K frames, 420K QAs & Perception, Prediction \\
\textbf{DriveMLLM \cite{guo2024drivemllm}} & 2024 & nuScenes & QA & 880 frames, 4,666 QAs & Perception \\
\textbf{nuPrompt \cite{wu2025language}} & 2025 & nuScenes & Language prompts & 850 videos, 35,367 prompts & Perception \\
\textbf{DriveLMM-o1 \cite{ishaq2025drivelmm}} & 2025 & nuScenes & Reasoning process & 1,962 frames, 18K QAs & Perception, Planning \\
\textbf{BDD-OIA \cite{xu2020explainable}} & 2020 & BDD100K & Explanation & 22,924 clips, 35,366 explanations & Perception \\
\textbf{BDD-X \cite{kim2018textual}} & 2018 & BDD100K & Description, Explanation & 6,984 clips, 50,298 description & Planning \\
\textbf{DriveGPT4 \cite{xu2024drivegpt4}} & 2024 & BDD-X & QA & 16K QAs, 40K conversations & Planning \\
\textbf{Refer-KITTI \cite{wu2023referring}} & 2023 & KITTI & Language prompts & 18 videos, 818 expressions & Perception \\
\textbf{WOMD-Reasoning \cite{li2024womd}} & 2024 & WOMD & QA, Prediction & 63k scenes, 3M QAs  & Prediction \\
\textbf{CityFlow-NL \cite{feng2021cityflow}} & 2021 & CityFlow & Language prompts & 3,028 tracks, 5,289 prompts & Perception \\
\textbf{DRIVINGVQA \cite{corbiere2025drivingvqa}} & 2025 & Code de la Route & VQA & 3,142 frames, 3142 QAs & Perception \\
\textbf{Highway-Text \cite{liao2025cot}} & 2025 & NGSIM, HighD & Language prompts & 6,606 scenarios & Prediction \\
\textbf{Urban-Text \cite{liao2025cot}} & 2025 & MoCAD, ApolloScape & Language prompts & 5,431 samples & Prediction \\
\textbf{MAPLM \cite{cao2024maplm}} & 2024 & THMA & Language prompts & 2M scenarios, 2M prompts & Perception \\
\textbf{MAPLM-QA \cite{cao2024maplm}} & 2024 & THMA & QA & 14K scenarios, 61K QAs & Perception \\
\textbf{Rank2Tell \cite{sachdeva2024rank2tell}} & 2024 & Self-collected & Chain VQA & 116 20-second clips & Perception \\
\textbf{DRAMA \cite{malla2023drama}} & 2023 & Self-collected & Chain QA & 17,785 scenarios, 103K QAs & Planning \\
\textbf{SUP-AD \cite{tian2024drivevlm}} & 2024 & Self-collected & Language prompts & 1,000 clips, 40+ categories & Planning \\
\textbf{LingoQA \cite{marcu2024lingoqa}} & 2024 & Self-collected & VQA & 28K videos, 419K annotations & Perception \\
\textbf{DriveMLM \cite{wang2023drivemlm}} & 2023 & CARLA & Decision, Explanation & 50K routes, 30 scenarios & End-to-end \\
\textbf{LMDrive \cite{shao2024lmdrive}} & 2024 & CARLA & Navigation instructions & 64K clips, 464K instructions & End-to-end \\
\textbf{DriveLM \cite{sima2024drivelm}} & 2024 & CARLA, nuScenes & Graph VQA & 4,063 frames, 377K QAs & End-to-end \\
\textbf{DriveBench \cite{xie2025vlms}} & 2025 & DriveLM & Graph VQA & 19,200 frames, 20,498 QAs & End-to-end \\
\textbf{DriveCoT \cite{wang2024drivecot}} & 2024 & CARLA & Reasoning process & 1,058 scenarios, 36K samples & End-to-end \\

\bottomrule
\end{tabular}
\end{table*}

\subsection{Cognition-augmented autonomous driving datasets}
The mainstream autonomous driving datasets typically combine real-world images, videos, and complementary sensor inputs from LiDAR, Radar, GPS, and IMU, providing annotations for various driving scenarios. These fundamental autonomous driving datasets primarily establish the mapping relationship between sensor perception data and core autonomous driving tasks, spanning perception, prediction, and planning. KITTI stands as a pioneering dataset that offers annotations for fundamental tasks such as object detection, traffic flow estimation, depth prediction, and object tracking \cite{geiger2013vision}. BDD100K contains more than 10,000 hours of driving video data, including image-level labels, object bounding boxes, drivable areas, lane markings, and full-frame instance segmentation annotations \cite{yu2020bdd100k}. As one of the most widely used public datasets, nuScenes provides 1,000 carefully curated urban driving scenes from Boston and Singapore, with a strong focus on addressing real-world long-tail distribution challenges \cite{caesar2020nuscenes}. The Waymo Open Dataset (WOMD) includes two sub-datasets: the perception dataset and the motion dataset, offering high-resolution sensor data, annotations, object trajectories, and corresponding 3D map information \cite{sun2020scalability}. HighD is a commonly adopted highway dataset for trajectory prediction. The data is smoothed, has low noise, and is of high quality \cite{krajewski2018highd}. NGSIM specializes in traffic flow analysis, with a focus on congested urban scenarios characterized by low vehicle speeds and intricate inter-vehicle interactions \cite{coifman2017critical}. CityFlow is used to solve vehicle re-identification and pedestrian detection problems \cite{tang2019cityflow}.

With the development of autonomous driving technology, there is an increasing demand for a deeper understanding of complex environments and improved model interpretability \cite{li2025explainable}. This has driven researchers to urgently build new datasets and benchmarks to support the development and validation of relevant systems. Concurrently, the widespread application of LLMs and CoT has introduced new dimensions to autonomous driving datasets \cite{liu2024survey}. Enhanced datasets are increasingly embedding cognition-augmented information, reflecting human-like decision-making processes, into existing multimodal foundational datasets. This integration further supports comprehensive scene understanding. Such datasets provide a solid conceptual basis for generating more coherent and human-like reasoning chains.

Research has demonstrated that incorporating specialized reasoning knowledge into the training data of LLMs can substantially improve their CoT reasoning ability \cite{sima2024drivelm}. Smaller models finetuned on datasets with reasoning data can elevate their accuracy to levels comparable to larger models, which have 1-2 times bigger parameter counts \cite{lobo2024impact}. Building upon this understanding of existing foundational datasets, this section aims to systematically analyze the current progress of cognition-augmented autonomous driving datasets.

To incorporate cognitive knowledge into datasets, a typical technical approach involves incremental annotation based on existing datasets. For instance, Talk2Car, as the first autonomous driving dataset based on natural language prompts, was built upon the nuScenes dataset \cite{deruyttere2019talk2car} \cite{caesar2020nuscenes}. Its annotations include selected keyframes focusing on individual specific objects within single images. To overcome the single-object annotation constraint, NuPrompt further expanded to associate multiple targets by providing textual descriptions of the positions and states of surrounding objects, thereby offering scene understanding information and achieving improvements in 3D driving scenarios \cite{wu2025language}.

Beyond semantic prompt annotation, the construction of driving datasets incorporating QA pairs has become  a key paradigm for embedding reasoning information. Serving as a critical connection between perception and cognition, Visual Question Answering (VQA) benchmarks validate and optimize the system's understanding and reasoning in complex scenarios, and improve decision-making interpretability \cite{atakishiyev2023explaining}. NuScenes-QA established the first VQA benchmark for autonomous driving \cite{qian2024nuscenes}. The annotations include key objects that may impact driving decisions or summarize the motivations behind specific driving behaviors. Additionally, the dataset also includes multimodal, multi-frame temporal information encompassing dynamic objects and static environments. However, NuScenes-QA emphasizes partial perception reasoning with limited single-word responses, lacking comprehensive scene-level analysis and annotations required for complex reasoning. NuScenesMQA advances answer quality through structured full-sentence responses, offering richer semantic hierarchies \cite{inoue2024nuscenes}. Single-step VQA cannot adequately model the multi-stage reasoning process of human drivers. To bridge this gap, Reason2Drive constructs a chain of question-answer pairs covering perception, prediction, and reasoning, combining nuScenes, Waymo, and ONCE datasets, to support explainable complex decision-making \cite{nie2024reason2drive}. Talk2BEV, by integrating LLMs, VLMs, and Bird’s Eye View (BEV) representations, significantly improves reasoning performance without relying on additional training \cite{choudhary2024talk2bev} \cite{mao2021one}. To address the limitations of autonomous driving systems in complex spatial understanding, DriveMLLM introduces spatial reasoning question-answer tasks, including absolute spatial reasoning (e.g., object positioning) and relative position reasoning (e.g., relative distances or front-back relationships between objects) \cite{guo2024drivemllm}. These tasks are specifically designed to evaluate and optimize the spatial understanding capabilities of MLLMs in driving scenarios. DriveLMM-o1 overcomes the limitations of focusing solely on answer accuracy by synthesizing multimodal inputs and providing manually revised chains of thought, ensuring logical consistency in the reasoning process and transparency of intermediate stages \cite{ishaq2025drivelmm}.

In addition to extending the nuScenes dataset, researchers have also developed cognitively enhanced data on other foundational datasets. BDD-OIA is based on BDD100K and includes numerous specialized annotations for object detection \cite{xu2020explainable}. BDD-X supplements driving conditions and decision-making rationale through textual descriptions, with a particular emphasis on extreme weather conditions \cite{kim2018textual}. To alleviate the manual burden while capturing meaningful data, some studies have adopted large language models for preliminary annotation. DriveGPT4 combines object detection models with GPT technology to extend recognition tasks and text description functions on top of BDD-X \cite{xu2024drivegpt4}. Refer-KITTI expands the original KITTI dataset, focusing on multi-object tracking to adapt to time-varying scenes \cite{wu2023referring}. The WOMD dataset adopts a fully automated data curation pipeline, first developing a rule-based program to interpret trajectory and high-definition map data, then using GPT-4 to generate interaction analysis and organizing the results into question-answer format \cite{li2024womd}. To address potential quality issues with LLM-generated question-answer pairs, DRIVINGVQA extracts  from authentic French driving theory exam samples, accompanied by expert annotations of relevant entities to answer questions and provide explanations \cite{corbiere2025drivingvqa}. This results in a high-quality visual reasoning dataset grounded in real-world driving knowledge.

Apart from the aforementioned extensions of mainstream datasets, several research institutions and companies have collected and developed their own cognitive-enhanced datasets. The Honda DRAMA dataset innovatively presents a dual challenge of risk object localization and risk explanation \cite{malla2023drama}. It uses the logical format of "what-which-where-how-why" to understand the vehicle's reaction to a single object as a question-answer chain, achieving object-level VQA. Rank2Tell provides object-level question answering for the perception module \cite{sachdeva2024rank2tell}. LingoQA adopts a free-form question-answer mechanism, expanding the scope of reasoning in autonomous driving to include nine distinct abilities \cite{marcu2024lingoqa}. To address safety-critical challenges, the Li Auto SUP-AD dataset tackles safety-critical challenges by mining long-tail data and applying precise annotations \cite{tian2024drivevlm}.

Another significant strategy for constructing cognitive-enhanced datasets involves leveraging simulation environments for data collection, integrating cognitive enhancement elements during the data generation phase \cite{dosovitskiy2017carla} \cite{song2023synthetic}. DriveLM combines the nuScenes dataset and CARLA collected data, introducing GVQA to emulate the human-like multi-step reasoning process \cite{sima2024drivelm}. However, it is limited to normal  normal scenarios in nuScenes and CARLA. LMDrive uses expert agents with access to CARLA's privileged information to collect data from more challenging cases, such as high-speed driving and lane changes \cite{shao2024lmdrive}. DriveMLM collects data from challenging driving scenarios featured in the CARLA Leaderboard 2.0 benchmark \cite{wang2023drivemlm} \cite{li2024think2drive}. Nevertheless, it only offers single-step reasoning explanations for driving decisions, lacking complete thought chain reasoning \cite{wang2023drivemlm}. The DriveCoT dataset is specifically designed to support CoT reasoning, which not only provides sensor data and decision control information but also annotates the full reasoning process \cite{wang2024drivecot}. It establishes a robust benchmark for evaluating the accuracy and interpretability of multi-step reasoning models in autonomous driving.

In summary, the development of current cognitive-enhanced autonomous driving datasets shows an evolutionary trend from single-object to multi-object, from short answers to complete sentence structures, and from single-step reasoning to multi-step chained reasoning. Through the construction of these diversified and targeted datasets, CoT reasoning models can acquire richer training data, enhancing the autonomous driving system's semantic understanding of complex scenarios, revealing the causal logic behind driving decisions, and advancing autonomous driving technology from perceptual intelligence to cognitive intelligence.

\subsection{Evaluation Metrics}

A comprehensive evaluation of Chain-of-Thought in autonomous driving necessitates multi-dimensional quantitative metrics, encompassing both reasoning quality and task performance. These complementary metrics collectively assess not only how CoT enhances model capabilities but also whether such theoretical enhancements materialize in operational scenarios. Reasoning metrics capture the model's evolving cognitive processes under CoT, while performance metrics validate the real-world efficacy of these cognitive advancements in driving scenarios \cite{guo2019safe}.

Reasoning metrics focus on quantifying the optimization effect of CoT on the upstream reasoning process \cite{golovneva2022roscoe}. They examine whether the model can generate logically coherent and causally correct decision-making grounds through step-by-step reasoning. Works like Nuscenes-QA, Drama, and Talk2bev directly applied NLP metrics such as BLEU, CIDEr, and METEOR scores to evaluate reasoning outputs \cite{papineni2002bleu} \cite{vedantam2015cider} \cite{banerjee2005meteor}. However, these metrics primarily assess text generation quality from a global perspective, failing to account for causal relationships between reasoning steps or their alignment with final decisions in driving scenarios. As a result, their effectiveness in evaluating reasoning remains limited. 

RDA-Driver, on the other hand, explicitly examines the alignment between reasoning and driving decisions, with some examples of CoT reasoning inconsistencies with subsequent decisions \cite{huang2024making}. For instance, analyses of LLaVA's CoT scores (higher is better) versus trajectory prediction errors (lower is better) demonstrate mismatches between reasoning and planning outcomes \cite{liu2023visual}. Although the model correctly reasoned the current state of the scene, such as recognizing that a vehicle ahead is stationary, the decision-making process followed an incorrect plan, directing the car to move forward. Furthermore, LingoQA introduced Lingo-Judge, a new evaluation metric inspired by GPT-Judge. This learning-based text classifier evaluates answer correctness and correlates model outputs with human preferences, offering a more nuanced measure of reasoning quality in autonomous driving \cite{marcu2024lingoqa}.

\begin{figure*}[htbp]
\centering
\includegraphics[width=6.4in]{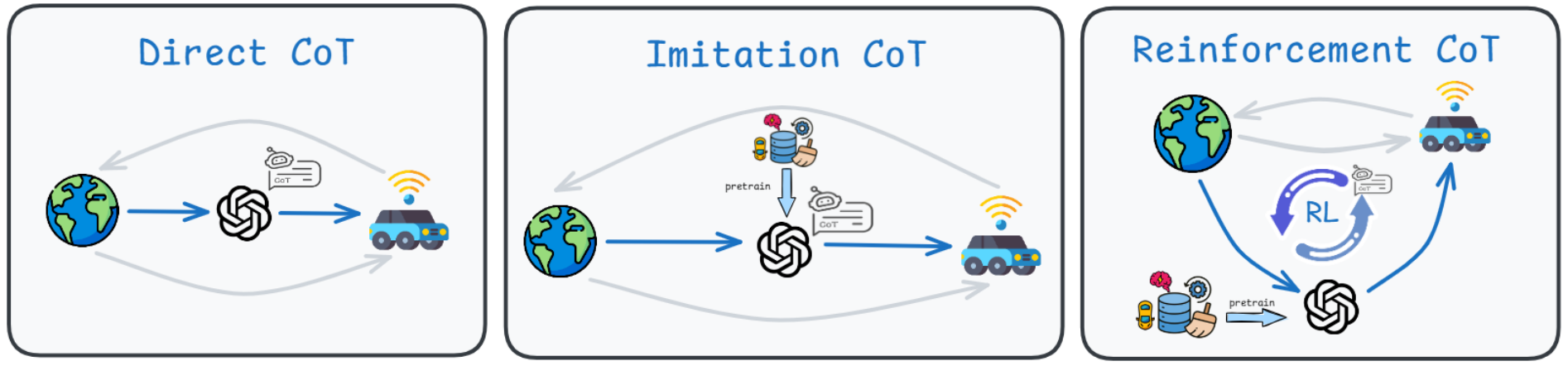}
\caption{Current autonomous driving research primarily employs Direct CoT and Imitation CoT for basic reasoning tasks, while emerging Reinforcement CoT approaches enable self-evolving reasoning capabilities that could lead to breakthrough "Aha Moments" in driving intelligence.}
\label{fig6} 
\end{figure*}

To better address the unique requirements of autonomous driving tasks and provide a more comprehensive evaluation framework, Reason2Drive introduces ADRScore \cite{nie2024reason2drive}. This benchmark is specifically designed to assess the quality of CoT reasoning in autonomous driving systems. It comprehensively evaluates autonomous driving reasoning alignment, redundancy, and missing steps, addressing the reasoning ambiguity issues present in existing metrics such as BLEU and CIDEr. Reasoning alignment (RA) measures the semantic similarity between the hypothesized reasoning and reference steps, redundancy (RD) detects unnecessary steps in the reasoning chain, and missing steps (MS) identifies critical missing reasoning steps. Additionally, ADRScore is extended to ADRScore-S, which adapts to tasks that include visual elements and uses the mean squared error of perception elements to rigorously assess the quality of visual reasoning. Overall, ADRScore provides a comprehensive and accurate method to evaluate the quality of reasoning chains in autonomous driving, particularly when handling complex spatial reasoning and decision-making \cite{xing2024autotrust}. This is a newly designed aggregation evaluation metric. DriveLMM-o1, inspired by VRC-Bench, systematically evaluates the accuracy of logical explanations generated by the model and the final predictions from seven dimensions: risk assessment accuracy, traffic rule adherence, scene awareness and object understanding, relevance, and missing details \cite{ishaq2025drivelmm}.

Performance metrics quantify the ultimate enhancement in driving performance attributable to CoT. In prediction tasks, average displacement error (ADE) and final displacement error (FDE) are typically used to evaluate the enhancement of CoT in prediction accuracy \cite{liao2025cot}. For comprehensive closed-loop evaluation in CARLA simulations, three key metrics are employed: the Driving Score (DS = RC/(1-IP)), which combines route completion rate (RC) with weighted infraction penalties (IP); Route Completion (RC) measuring the percentage of the predefined route successfully navigated (0-100\%); and Time Consumption (TC) recording total operation time until completion or failure \cite{sima2024drivelm} \cite{dosovitskiy2017carla}. In open-loop testing scenarios, performance is additionally assessed through L2 trajectory error, which quantifies the deviation from ground truth trajectories at 1s, 2s, and 3s intervals, and collision rate (CR), which measures the frequency of incidents \cite{qian2024nuscenes}. Together, these metrics form a rigorous framework for evaluating the impact of CoT on autonomous driving performance.

This integrated evaluation framework, combining both reasoning metrics and performance metrics, enables a comprehensive assessment of CoT-guided autonomous driving systems. By simultaneously examining the system's cognitive reasoning capability and its operational effectiveness in driving scenarios, this approach establishes a robust scientific basis for the systematic optimization of autonomous driving.

\section{Discussion}
In this section, we propose that the application of Chain-of-Thought (CoT) in autonomous driving has evolved through three distinct stages.
The first stage is Direct CoT, which leverages the intrinsic reasoning capabilities of foundational large models without requiring additional training or fine-tuning \cite{xing2025openemma}. By designing specific prompts, a fixed reasoning workflow is established for driving tasks \cite{wen2023dilu} \cite{jiang2024koma}. At each step of reasoning, domain-specific knowledge and typical examples are systematically incorporated, enabling the system to reason step-by-step before generating the final command. 

The next two stages involve fine-tuning parameters to endow the model with deeper reasoning abilities \cite{lobo2024impact}. The second stage, Imitation CoT, involves supervised fine-tuning (SFT) of pre-trained models on driving datasets \cite{wang2024drivecot} \cite{zhang2024wisead}. First, human thought semantics or the explicit reasoning chains from the first method are recorded, preserving the thought trajectories during driving to form an enhanced autonomous driving dataset, as described in Chapter IV \cite{wang2023drivemlm} \cite{shao2024lmdrive} \cite{sima2024drivelm}. Triplets such as <driving scenario, reasoning chain, driving action> are used as labeled data to fine-tune the large model, enabling it to replicate high-quality reasoning processes in similar scenarios. 

The third stage is Reinforcement CoT, which draws inspiration from the technical approach of the language model DeepSeek-R1, combining reinforcement learning (RL) with reasoning techniques. During GRPO training, DeepSeek-R1 experienced an "Aha Moment", achieving the emergence of long reasoning chains while significantly reducing data requirements \cite{guo2025deepseek}. This demonstrates the potential of RL in transitioning models from imitation intelligence to emergent intelligence. Reinforcement CoT adopts a two-step training process \cite{jiang2025alphadrive} \cite{qian2025agentthink}. In Step One, it resembles Imitation CoT, employing supervised fine-tuning for knowledge transfer and establishing a preliminary reasoning foundation. In Step Two, reinforcement learning is introduced to enable policy self-evolution, building upon the supervised fine-tuning. Through trial-and-error and reward-based guidance, the model's outputs are refined, mitigating instability from Step One. The system autonomously identifies reasoning flaws and seeks optimized pathways, forming a self-reflective reasoning chain that may trigger an "Aha Moment" in autonomous driving.  

Currently, most research relies on the first two stages, Direct CoT and Imitation CoT. These methods primarily leverage pre-trained models and simple SFT, with limited exploration into further optimization of reasoning strategies. A few emerging studies have introduced Reinforcement CoT, combining SFT and RL to develop autonomous driving reasoning frameworks that promote model self-evolution.

\section{Challenges and Future Directions}

\subsection{Challenges for CoT in autonomous driving}

While the application of CoT reasoning in autonomous driving is  promising, it still faces several critical challenges in practice. Based on the review of current research, these challenges can be summarized across four key dimensions:

\textbf{Cross-modal gap:}
Under the CoT framework, autonomous driving systems typically rely on multiple modalities such as vision, language, and motion. However, discrepancies exist between visual and semantic features at the perceptual level, leading to inter-modal semantic deviations in scene understanding \cite{huang2024making}. In complex driving scenarios, current models struggle to accurately map key image details to high-quality linguistic descriptions. Furthermore, in the reasoning chain from semantic comprehension to action execution, multi-step reasoning errors accumulate as CoT unfolds, resulting in significant deviations in final actions \cite{kim2024openvla}. Bridging the cross-modal gap represents a fundamental requirement for developing trustworthy autonomous systems.

\textbf{Alignment with human cognition:}
CoT attempts to simulate human thought processes to progressively accomplish reasoning tasks \cite{li2024large}. However, human reasoning relies on commonsense knowledge, domain expertise, and contextual understanding. These factors are difficult to model through simple rules or purely data-driven methods, leading to misalignment between generated reasoning chains and human cognition \cite{song2024preference} \cite{kothawade2021auto}. Ensuring that CoT reasoning aligns with human thought processes to enhance system interpretability and reliability remains an urgent problem to address.
\textbf{Balancing reasoning depth and speed:}
CoT reasoning depends on a series of intermediate steps, which are at odds with the stringent real-time requirements of autonomous driving. Current large-scale pre-trained models suffer from high latency and substantial computational costs, rendering them impractical for deployment on embedded, resource-constrained platforms. Compressing the reasoning chain and computational load, without compromising decision quality, coheres as a persistent obstacle for CoT deployment \cite{xu2025chain}.

\textbf{Safety verification:}
Despite its enhanced interpretability, CoT reasoning introduces uncertainty through extended reasoning chains. In corner cases, systems may produce logically coherent but factually incorrect outputs, known as large model hallucination, resulting in misleading decisions \cite{ji2023survey}. In safety-critical tasks like autonomous driving, even minor errors can have severe consequences. The absence of robust mechanisms for dynamic risk evaluation and intervention within the CoT process further complicates reliable deployment.

\subsection{Future Directions}
\textbf{Interference Mechanism:}
In complex reasoning tasks for autonomous driving, the lack of self-correction mechanisms may lead to the gradual amplification of single-step errors in chained reasoning structures. It is advisable to introduce "interference mechanisms" during both training and testing phases. During training, "adversarial QA interference samples" can be incorporated by systematically constructing training cases with prompt biases, semantic ambiguities, and noise interference to enhance the model's ability to recognize and handle abnormal inputs. Concurrently, dynamic interference testing should be conducted during the validation phase to evaluate the model's anti-interference performance in simulated driving scenarios.

\textbf{Advanced CoT:}
Currently, the autonomous driving field primarily relies on Chain-of-Thought for multi-step reasoning, but its computational overhead is substantial. Recent studies have proposed explicitly compact CoT variants, such as Chain-of-Draft, which significantly improve reasoning efficiency \cite{xu2025chain} \cite{manas2024cot}. Additionally, implicitly latent CoT variants achieve low-latency responses while maintaining decision-making accuracy by substantially reducing textual volume and employing latent states to compress high-level semantic information, making them suitable for autonomous driving applications \cite{deng2024explicit} \cite{kong2025scalable}.

\textbf{Collaborative Fast and Slow Thinking:}
Inspired by research on human neural mechanisms, some studies adopt Collaborative Slow and Fast-thinking Systems, which combine rapid intuitive reasoning with in-depth logical analysis \cite{li2025system}. This approach dynamically allocates computational resources in autonomous driving systems to achieve a balance between reasoning depth and real-time performance.

\section{Conclusion}
This survey provides a comprehensive overview of Chain-of-Thought reasoning in autonomous driving. It encompasses the background, method classification, relevant datasets, key challenges, and future directions. The article thoroughly reviews and analyzes the application of existing CoT in various tasks of autonomous driving, including perception understanding, prediction planning, decision-making control, and end-to-end autonomous driving. The paper further categorizes CoT-based reasoning in autonomous driving according to model structures into modular CoT, logical CoT, and reflective CoT. It summarizes the available reasoning driving datasets in the field so far and elaborates on the advantages of incorporating CoT into the domain. Based on current research, the paper also discusses existing and future challenges. Some potential solutions and future exploration directions are also proposed, such as combining CoT with self-learning methods. The aim is to spotlight CoT in autonomous driving and foster deeper research in this field.
\ifCLASSOPTIONcaptionsoff
  \newpage
\fi

\bibliographystyle{IEEEtran}
\bibliography{reference.bib}

\begin{thebibliography}{100}
\providecommand{\url}[1]{#1}
\csname url@samestyle\endcsname
\providecommand{\newblock}{\relax}
\providecommand{\bibinfo}[2]{#2}
\providecommand{\BIBentrySTDinterwordspacing}{\spaceskip=0pt\relax}
\providecommand{\BIBentryALTinterwordstretchfactor}{4}
\providecommand{\BIBentryALTinterwordspacing}{\spaceskip=\fontdimen2\font plus
\BIBentryALTinterwordstretchfactor\fontdimen3\font minus \fontdimen4\font\relax}
\providecommand{\BIBforeignlanguage}[2]{{%
\expandafter\ifx\csname l@#1\endcsname\relax
\typeout{** WARNING: IEEEtran.bst: No hyphenation pattern has been}%
\typeout{** loaded for the language `#1'. Using the pattern for}%
\typeout{** the default language instead.}%
\else
\language=\csname l@#1\endcsname
\fi
#2}}
\providecommand{\BIBdecl}{\relax}
\BIBdecl

\bibitem{vaswani2017attention}
A.~Vaswani, N.~Shazeer, N.~Parmar, J.~Uszkoreit, L.~Jones, A.~N. Gomez, {\L}.~Kaiser, and I.~Polosukhin, ``Attention is all you need,'' \emph{Advances in neural information processing systems}, vol.~30, 2017.

\bibitem{brown2020language}
T.~Brown, B.~Mann, N.~Ryder, M.~Subbiah, J.~D. Kaplan, P.~Dhariwal, A.~Neelakantan, P.~Shyam, G.~Sastry, A.~Askell \emph{et~al.}, ``Language models are few-shot learners,'' \emph{Advances in neural information processing systems}, vol.~33, pp. 1877--1901, 2020.

\bibitem{liu2023pre}
P.~Liu, W.~Yuan, J.~Fu, Z.~Jiang, H.~Hayashi, and G.~Neubig, ``Pre-train, prompt, and predict: A systematic survey of prompting methods in natural language processing,'' \emph{ACM computing surveys}, vol.~55, no.~9, pp. 1--35, 2023.

\bibitem{touvron2023llama}
H.~Touvron, T.~Lavril, G.~Izacard, X.~Martinet, M.-A. Lachaux, T.~Lacroix, B.~Rozi{\`e}re, N.~Goyal, E.~Hambro, F.~Azhar \emph{et~al.}, ``Llama: Open and efficient foundation language models,'' \emph{arXiv preprint arXiv:2302.13971}, 2023.

\bibitem{zhao2023brain}
L.~Zhao, L.~Zhang, Z.~Wu, Y.~Chen, H.~Dai, X.~Yu, Z.~Liu, T.~Zhang, X.~Hu, X.~Jiang \emph{et~al.}, ``When brain-inspired ai meets agi,'' \emph{Meta-Radiology}, vol.~1, no.~1, p. 100005, 2023.

\bibitem{kiran2021deep}
B.~R. Kiran, I.~Sobh, V.~Talpaert, P.~Mannion, A.~A. Al~Sallab, S.~Yogamani, and P.~P{\'e}rez, ``Deep reinforcement learning for autonomous driving: A survey,'' \emph{IEEE Transactions on Intelligent Transportation Systems}, vol.~23, no.~6, pp. 4909--4926, 2021.

\bibitem{wu2024prospective}
J.~Wu, B.~Gao, J.~Gao, J.~Yu, H.~Chu, Q.~Yu, X.~Gong, Y.~Chang, H.~E. Tseng, H.~Chen \emph{et~al.}, ``Prospective role of foundation models in advancing autonomous vehicles,'' \emph{Research}, vol.~7, p. 0399, 2024.

\bibitem{cui2024large}
C.~Cui, Y.~Ma, Z.~Yang, Y.~Zhou, P.~Liu, J.~Lu, L.~Li, Y.~Chen, J.~H. Panchal, A.~Abdelraouf \emph{et~al.}, ``Large language models for autonomous driving (llm4ad): Concept, benchmark, simulation, and real-vehicle experiment,'' \emph{arXiv preprint arXiv:2410.15281}, 2024.

\bibitem{cui2024survey}
C.~Cui, Y.~Ma, X.~Cao, W.~Ye, Y.~Zhou, K.~Liang, J.~Chen, J.~Lu, Z.~Yang, K.-D. Liao \emph{et~al.}, ``A survey on multimodal large language models for autonomous driving,'' in \emph{Proceedings of the IEEE/CVF Winter Conference on Applications of Computer Vision}, 2024, pp. 958--979.

\bibitem{zhao2024survey}
R.~Zhao, Y.~Li, Y.~Fan, F.~Gao, M.~Tsukada, and Z.~Gao, ``A survey on recent advancements in autonomous driving using deep reinforcement learning: Applications, challenges, and solutions,'' \emph{IEEE Transactions on Intelligent Transportation Systems}, 2024.

\bibitem{chen2024end}
L.~Chen, P.~Wu, K.~Chitta, B.~Jaeger, A.~Geiger, and H.~Li, ``End-to-end autonomous driving: Challenges and frontiers,'' \emph{IEEE Transactions on Pattern Analysis and Machine Intelligence}, 2024.

\bibitem{huang2022towards}
J.~Huang and K.~C.-C. Chang, ``Towards reasoning in large language models: A survey,'' \emph{arXiv preprint arXiv:2212.10403}, 2022.

\bibitem{wei2022chain}
J.~Wei, X.~Wang, D.~Schuurmans, M.~Bosma, F.~Xia, E.~Chi, Q.~V. Le, D.~Zhou \emph{et~al.}, ``Chain-of-thought prompting elicits reasoning in large language models,'' \emph{Advances in neural information processing systems}, vol.~35, pp. 24\,824--24\,837, 2022.

\bibitem{xiang2025towards}
V.~Xiang, C.~Snell, K.~Gandhi, A.~Albalak, A.~Singh, C.~Blagden, D.~Phung, R.~Rafailov, N.~Lile, D.~Mahan \emph{et~al.}, ``Towards system 2 reasoning in llms: Learning how to think with meta chain-of-though,'' \emph{arXiv preprint arXiv:2501.04682}, 2025.

\bibitem{jaech2024openai}
A.~Jaech, A.~Kalai, A.~Lerer, A.~Richardson, A.~El-Kishky, A.~Low, A.~Helyar, A.~Madry, A.~Beutel, A.~Carney \emph{et~al.}, ``Openai o1 system card,'' \emph{arXiv preprint arXiv:2412.16720}, 2024.

\bibitem{guo2025deepseek}
D.~Guo, D.~Yang, H.~Zhang, J.~Song, R.~Zhang, R.~Xu, Q.~Zhu, S.~Ma, P.~Wang, X.~Bi \emph{et~al.}, ``Deepseek-r1: Incentivizing reasoning capability in llms via reinforcement learning,'' \emph{arXiv preprint arXiv:2501.12948}, 2025.

\bibitem{manas2024cot}
K.~Manas, S.~Zwicklbauer, and A.~Paschke, ``Cot-tl: Low-resource temporal knowledge representation of planning instructions using chain-of-thought reasoning,'' in \emph{2024 IEEE/RSJ International Conference on Intelligent Robots and Systems (IROS)}.\hskip 1em plus 0.5em minus 0.4em\relax IEEE, 2024, pp. 9636--9643.

\bibitem{wang2024drivecot}
T.~Wang, E.~Xie, R.~Chu, Z.~Li, and P.~Luo, ``Drivecot: Integrating chain-of-thought reasoning with end-to-end driving,'' \emph{arXiv preprint arXiv:2403.16996}, 2024.

\bibitem{liu2024medcot}
J.~Liu, Y.~Wang, J.~Du, J.~T. Zhou, and Z.~Liu, ``Medcot: Medical chain of thought via hierarchical expert,'' \emph{arXiv preprint arXiv:2412.13736}, 2024.

\bibitem{deng2024leveraging}
Y.~Deng, X.~Zhang, D.~Zhou, D.~Zhang, and B.~Huang, ``Leveraging nlp in finance: A synergistic approach using large language models and chain-of-thought reasoning,'' in \emph{Proceedings of the 5th International Conference on Artificial Intelligence and Computer Engineering}, 2024, pp. 494--500.

\bibitem{chen2024m}
Q.~Chen, L.~Qin, J.~Zhang, Z.~Chen, X.~Xu, and W.~Che, ``M$^{3}$ cot: A novel benchmark for multi-domain multi-step multi-modal chain-of-thought,'' \emph{arXiv preprint arXiv:2405.16473}, 2024.

\bibitem{fu2024drive}
D.~Fu, X.~Li, L.~Wen, M.~Dou, P.~Cai, B.~Shi, and Y.~Qiao, ``Drive like a human: Rethinking autonomous driving with large language models,'' in \emph{2024 IEEE/CVF Winter Conference on Applications of Computer Vision Workshops (WACVW)}.\hskip 1em plus 0.5em minus 0.4em\relax IEEE, 2024, pp. 910--919.

\bibitem{ma2024learning}
Y.~Ma, X.~Cao, W.~Ye, C.~Cui, K.~Mei, and Z.~Wang, ``Learning autonomous driving tasks via human feedbacks with large language models,'' in \emph{Findings of the Association for Computational Linguistics: EMNLP 2024}, 2024, pp. 4985--4995.

\bibitem{tian2024drivevlm}
X.~Tian, J.~Gu, B.~Li, Y.~Liu, Y.~Wang, Z.~Zhao, K.~Zhan, P.~Jia, X.~Lang, and H.~Zhao, ``Drivevlm: The convergence of autonomous driving and large vision-language models,'' \emph{arXiv preprint arXiv:2402.12289}, 2024.

\bibitem{nie2024reason2drive}
M.~Nie, R.~Peng, C.~Wang, X.~Cai, J.~Han, H.~Xu, and L.~Zhang, ``Reason2drive: Towards interpretable and chain-based reasoning for autonomous driving,'' in \emph{European Conference on Computer Vision}.\hskip 1em plus 0.5em minus 0.4em\relax Springer, 2024, pp. 292--308.

\bibitem{yang2023llm4drive}
Z.~Yang, X.~Jia, H.~Li, and J.~Yan, ``Llm4drive: A survey of large language models for autonomous driving,'' \emph{arXiv preprint arXiv:2311.01043}, 2023.

\bibitem{gao2024survey}
H.~Gao, Z.~Wang, Y.~Li, K.~Long, M.~Yang, and Y.~Shen, ``A survey for foundation models in autonomous driving,'' \emph{arXiv preprint arXiv:2402.01105}, 2024.

\bibitem{zhou2024vision}
X.~Zhou, M.~Liu, E.~Yurtsever, B.~L. Zagar, W.~Zimmer, H.~Cao, and A.~C. Knoll, ``Vision language models in autonomous driving: A survey and outlook,'' \emph{IEEE Transactions on Intelligent Vehicles}, 2024.

\bibitem{chen2025towards}
Q.~Chen, L.~Qin, J.~Liu, D.~Peng, J.~Guan, P.~Wang, M.~Hu, Y.~Zhou, T.~Gao, and W.~Che, ``Towards reasoning era: A survey of long chain-of-thought for reasoning large language models,'' \emph{arXiv preprint arXiv:2503.09567}, 2025.

\bibitem{yu2023towards}
Z.~Yu, L.~He, Z.~Wu, X.~Dai, and J.~Chen, ``Towards better chain-of-thought prompting strategies: A survey,'' \emph{arXiv preprint arXiv:2310.04959}, 2023.

\bibitem{o2018scalable}
M.~O'Kelly, A.~Sinha, H.~Namkoong, R.~Tedrake, and J.~C. Duchi, ``Scalable end-to-end autonomous vehicle testing via rare-event simulation,'' \emph{Advances in neural information processing systems}, vol.~31, 2018.

\bibitem{hu2023planning}
Y.~Hu, J.~Yang, L.~Chen, K.~Li, C.~Sima, X.~Zhu, S.~Chai, S.~Du, T.~Lin, W.~Wang \emph{et~al.}, ``Planning-oriented autonomous driving,'' in \emph{Proceedings of the IEEE/CVF conference on computer vision and pattern recognition}, 2023, pp. 17\,853--17\,862.

\bibitem{zhang2021end}
Z.~Zhang, A.~Liniger, D.~Dai, F.~Yu, and L.~Van~Gool, ``End-to-end urban driving by imitating a reinforcement learning coach,'' in \emph{Proceedings of the IEEE/CVF international conference on computer vision}, 2021, pp. 15\,222--15\,232.

\bibitem{pan2020imitation}
Y.~Pan, C.-A. Cheng, K.~Saigol, K.~Lee, X.~Yan, E.~A. Theodorou, and B.~Boots, ``Imitation learning for agile autonomous driving,'' \emph{The International Journal of Robotics Research}, vol.~39, no. 2-3, pp. 286--302, 2020.

\bibitem{cui2025sustainable}
Y.~Cui, S.~Yang, C.~Wan, X.~Li, J.~Xing, Y.~Zhang, Y.~Huang, and H.~Chen, ``Sustainable adaptation for autonomous driving with the mixture of progressive experts networ,'' \emph{arXiv preprint arXiv:2502.05943}, 2025.

\bibitem{li2023towards}
X.~Li, Y.~Bai, P.~Cai, L.~Wen, D.~Fu, B.~Zhang, X.~Yang, X.~Cai, T.~Ma, J.~Guo \emph{et~al.}, ``Towards knowledge-driven autonomous driving,'' \emph{arXiv preprint arXiv:2312.04316}, 2023.

\bibitem{treiber2000congested}
M.~Treiber, A.~Hennecke, and D.~Helbing, ``Congested traffic states in empirical observations and microscopic simulations,'' \emph{Physical review E}, vol.~62, no.~2, p. 1805, 2000.

\bibitem{xiao2021rule}
W.~Xiao, N.~Mehdipour, A.~Collin, A.~Y. Bin-Nun, E.~Frazzoli, R.~D. Tebbens, and C.~Belta, ``Rule-based optimal control for autonomous driving,'' in \emph{Proceedings of the ACM/IEEE 12th International Conference on Cyber-Physical Systems}, 2021, pp. 143--154.

\bibitem{shan2020reinforcement}
Y.~Shan, B.~Zheng, L.~Chen, L.~Chen, and D.~Chen, ``A reinforcement learning-based adaptive path tracking approach for autonomous driving,'' \emph{IEEE Transactions on Vehicular Technology}, vol.~69, no.~10, pp. 10\,581--10\,595, 2020.

\bibitem{wang2022high}
L.~Wang, C.~Fernandez, and C.~Stiller, ``High-level decision making for automated highway driving via behavior cloning,'' \emph{IEEE Transactions on Intelligent Vehicles}, vol.~8, no.~1, pp. 923--935, 2022.

\bibitem{wen2023dilu}
L.~Wen, D.~Fu, X.~Li, X.~Cai, T.~Ma, P.~Cai, M.~Dou, B.~Shi, L.~He, and Y.~Qiao, ``Dilu: A knowledge-driven approach to autonomous driving with large language models,'' \emph{arXiv preprint arXiv:2309.16292}, 2023.

\bibitem{zhang2024wisead}
S.~Zhang, W.~Huang, Z.~Gao, H.~Chen, and C.~Lv, ``Wisead: Knowledge augmented end-to-end autonomous driving with vision-language model,'' \emph{arXiv preprint arXiv:2412.09951}, 2024.

\bibitem{brachman2004knowledge}
R.~Brachman and H.~Levesque, \emph{Knowledge representation and reasoning}.\hskip 1em plus 0.5em minus 0.4em\relax Elsevier, 2004.

\bibitem{wang2022towards}
B.~Wang, S.~Min, X.~Deng, J.~Shen, Y.~Wu, L.~Zettlemoyer, and H.~Sun, ``Towards understanding chain-of-thought prompting: An empirical study of what matters,'' \emph{arXiv preprint arXiv:2212.10001}, 2022.

\bibitem{xing2024comprehensive}
J.~Xing, D.~Wei, S.~Zhou, T.~Wang, Y.~Huang, and H.~Chen, ``A comprehensive study on self-learning methods and implications to autonomous driving,'' \emph{IEEE Transactions on Neural Networks and Learning Systems}, 2024.

\bibitem{tu2025role}
S.~Tu, X.~Zhou, D.~Liang, X.~Jiang, Y.~Zhang, X.~Li, and X.~Bai, ``The role of world models in shaping autonomous driving: A comprehensive survey,'' \emph{arXiv preprint arXiv:2502.10498}, 2025.

\bibitem{pathak2017curiosity}
D.~Pathak, P.~Agrawal, A.~A. Efros, and T.~Darrell, ``Curiosity-driven exploration by self-supervised prediction,'' in \emph{International conference on machine learning}.\hskip 1em plus 0.5em minus 0.4em\relax PMLR, 2017, pp. 2778--2787.

\bibitem{zhang2025enhancing}
X.~Zhang, K.~Wang, T.~Hu, and H.~Ma, ``Enhancing autonomous driving through dual-process learning with behavior and reflection integration,'' in \emph{ICASSP 2025-2025 IEEE International Conference on Acoustics, Speech and Signal Processing (ICASSP)}.\hskip 1em plus 0.5em minus 0.4em\relax IEEE, 2025, pp. 1--5.

\bibitem{mei2024continuously}
J.~Mei, Y.~Ma, X.~Yang, L.~Wen, X.~Cai, X.~Li, D.~Fu, B.~Zhang, P.~Cai, M.~Dou \emph{et~al.}, ``Continuously learning, adapting, and improving: A dual-process approach to autonomous driving,'' \emph{arXiv preprint arXiv:2405.15324}, 2024.

\bibitem{team2023gemini}
G.~Team, R.~Anil, S.~Borgeaud, J.-B. Alayrac, J.~Yu, R.~Soricut, J.~Schalkwyk, A.~M. Dai, A.~Hauth, K.~Millican \emph{et~al.}, ``Gemini: a family of highly capable multimodal models,'' \emph{arXiv preprint arXiv:2312.11805}, 2023.

\bibitem{hurst2024gpt}
A.~Hurst, A.~Lerer, A.~P. Goucher, A.~Perelman, A.~Ramesh, A.~Clark, A.~Ostrow, A.~Welihinda, A.~Hayes, A.~Radford \emph{et~al.}, ``Gpt-4o system card,'' \emph{arXiv preprint arXiv:2410.21276}, 2024.

\bibitem{liu2024deepseek}
A.~Liu, B.~Feng, B.~Xue, B.~Wang, B.~Wu, C.~Lu, C.~Zhao, C.~Deng, C.~Zhang, C.~Ruan \emph{et~al.}, ``Deepseek-v3 technical report,'' \emph{arXiv preprint arXiv:2412.19437}, 2024.

\bibitem{li2024enhancing}
Z.~Li, D.~Liu, C.~Zhang, H.~Wang, T.~Xue, and W.~Cai, ``Enhancing advanced visual reasoning ability of large language models,'' \emph{arXiv preprint arXiv:2409.13980}, 2024.

\bibitem{zhang2024mm}
D.~Zhang, Y.~Yu, J.~Dong, C.~Li, D.~Su, C.~Chu, and D.~Yu, ``Mm-llms: Recent advances in multimodal large language models,'' \emph{arXiv preprint arXiv:2401.13601}, 2024.

\bibitem{li2025visual}
Y.~Li, Z.~Lai, W.~Bao, Z.~Tan, A.~Dao, K.~Sui, J.~Shen, D.~Liu, H.~Liu, and Y.~Kong, ``Visual large language models for generalized and specialized applications,'' \emph{arXiv preprint arXiv:2501.02765}, 2025.

\bibitem{zhou2025opendrivevla}
X.~Zhou, X.~Han, F.~Yang, Y.~Ma, and A.~C. Knoll, ``Opendrivevla: Towards end-to-end autonomous driving with large vision language action model,'' \emph{arXiv preprint arXiv:2503.23463}, 2025.

\bibitem{jiang2025alphadrive}
B.~Jiang, S.~Chen, Q.~Zhang, W.~Liu, and X.~Wang, ``Alphadrive: Unleashing the power of vlms in autonomous driving via reinforcement learning and reasoning,'' \emph{arXiv preprint arXiv:2503.07608}, 2025.

\bibitem{qu2025survey}
X.~Qu, Y.~Li, Z.~Su, W.~Sun, J.~Yan, D.~Liu, G.~Cui, D.~Liu, S.~Liang, J.~He \emph{et~al.}, ``A survey of efficient reasoning for large reasoning models: Language, multimodality, and beyond,'' \emph{arXiv preprint arXiv:2503.21614}, 2025.

\bibitem{dasgupta2022language}
I.~Dasgupta, A.~K. Lampinen, S.~C. Chan, H.~R. Sheahan, A.~Creswell, D.~Kumaran, J.~L. McClelland, and F.~Hill, ``Language models show human-like content effects on reasoning tasks,'' \emph{arXiv preprint arXiv:2207.07051}, 2022.

\bibitem{sprague2024cot}
Z.~Sprague, F.~Yin, J.~D. Rodriguez, D.~Jiang, M.~Wadhwa, P.~Singhal, X.~Zhao, X.~Ye, K.~Mahowald, and G.~Durrett, ``To cot or not to cot? chain-of-thought helps mainly on math and symbolic reasoning,'' \emph{arXiv preprint arXiv:2409.12183}, 2024.

\bibitem{lightman2023let}
H.~Lightman, V.~Kosaraju, Y.~Burda, H.~Edwards, B.~Baker, T.~Lee, J.~Leike, J.~Schulman, I.~Sutskever, and K.~Cobbe, ``Let's verify step by step,'' in \emph{The Twelfth International Conference on Learning Representations}, 2023.

\bibitem{kojima2022large}
T.~Kojima, S.~S. Gu, M.~Reid, Y.~Matsuo, and Y.~Iwasawa, ``Large language models are zero-shot reasoners,'' \emph{Advances in neural information processing systems}, vol.~35, pp. 22\,199--22\,213, 2022.

\bibitem{shaikh2022second}
O.~Shaikh, H.~Zhang, W.~Held, M.~Bernstein, and D.~Yang, ``On second thought, let's not think step by step! bias and toxicity in zero-shot reasoning,'' \emph{arXiv preprint arXiv:2212.08061}, 2022.

\bibitem{zhang2024supervised}
X.~Zhang and D.~Ding, ``Supervised chain of thought,'' \emph{arXiv preprint arXiv:2410.14198}, 2024.

\bibitem{wang2022self}
X.~Wang, J.~Wei, D.~Schuurmans, Q.~Le, E.~Chi, S.~Narang, A.~Chowdhery, and D.~Zhou, ``Self-consistency improves chain of thought reasoning in language models,'' \emph{arXiv preprint arXiv:2203.11171}, 2022.

\bibitem{yao2023tree}
S.~Yao, D.~Yu, J.~Zhao, I.~Shafran, T.~Griffiths, Y.~Cao, and K.~Narasimhan, ``Tree of thoughts: Deliberate problem solving with large language models,'' \emph{Advances in neural information processing systems}, vol.~36, pp. 11\,809--11\,822, 2023.

\bibitem{besta2024graph}
M.~Besta, N.~Blach, A.~Kubicek, R.~Gerstenberger, M.~Podstawski, L.~Gianinazzi, J.~Gajda, T.~Lehmann, H.~Niewiadomski, P.~Nyczyk \emph{et~al.}, ``Graph of thoughts: Solving elaborate problems with large language models,'' in \emph{Proceedings of the AAAI Conference on Artificial Intelligence}, vol.~38, no.~16, 2024, pp. 17\,682--17\,690.

\bibitem{shao2024visual}
H.~Shao, S.~Qian, H.~Xiao, G.~Song, Z.~Zong, L.~Wang, Y.~Liu, and H.~Li, ``Visual cot: Advancing multi-modal language models with a comprehensive dataset and benchmark for chain-of-thought reasoning,'' \emph{Advances in Neural Information Processing Systems}, vol.~37, pp. 8612--8642, 2024.

\bibitem{zhang2023multimodal}
Z.~Zhang, A.~Zhang, M.~Li, H.~Zhao, G.~Karypis, and A.~Smola, ``Multimodal chain-of-thought reasoning in language models,'' \emph{arXiv preprint arXiv:2302.00923}, 2023.

\bibitem{mao2023language}
J.~Mao, J.~Ye, Y.~Qian, M.~Pavone, and Y.~Wang, ``A language agent for autonomous driving,'' \emph{arXiv preprint arXiv:2311.10813}, 2023.

\bibitem{ma2024dolphins}
Y.~Ma, Y.~Cao, J.~Sun, M.~Pavone, and C.~Xiao, ``Dolphins: Multimodal language model for driving,'' in \emph{European Conference on Computer Vision}.\hskip 1em plus 0.5em minus 0.4em\relax Springer, 2024, pp. 403--420.

\bibitem{corbière2025retrievalbasedinterleavedvisualchainofthought}
\BIBentryALTinterwordspacing
C.~Corbière, S.~Roburin, S.~Montariol, A.~Bosselut, and A.~Alahi, ``Retrieval-based interleaved visual chain-of-thought in real-world driving scenarios,'' 2025. [Online]. Available: \url{https://arxiv.org/abs/2501.04671}
\BIBentrySTDinterwordspacing

\bibitem{hou2025driveagent}
X.~Hou, W.~Wang, L.~Yang, H.~Lin, J.~Feng, H.~Min, and X.~Zhao, ``Driveagent: Multi-agent structured reasoning with llm and multimodal sensor fusion for autonomous driving,'' \emph{arXiv preprint arXiv:2505.02123}, 2025.

\bibitem{qian2025agentthink}
K.~Qian, S.~Jiang, Y.~Zhong, Z.~Luo, Z.~Huang, T.~Zhu, K.~Jiang, M.~Yang, Z.~Fu, J.~Miao \emph{et~al.}, ``Agentthink: A unified framework for tool-augmented chain-of-thought reasoning in vision-language models for autonomous driving,'' \emph{arXiv preprint arXiv:2505.15298}, 2025.

\bibitem{li2024womd}
Y.~Li, C.~Ge, C.~Li, C.~Xu, M.~Tomizuka, C.~Tang, M.~Ding, and W.~Zhan, ``Womd-reasoning: A large-scale language dataset for interaction and driving intentions reasoning,'' \emph{arXiv preprint arXiv:2407.04281}, 2024.

\bibitem{peng2025lc}
M.~Peng, X.~Guo, X.~Chen, K.~Chen, M.~Zhu, L.~Chen, and F.-Y. Wang, ``Lc-llm: Explainable lane-change intention and trajectory predictions with large language models,'' \emph{Communications in Transportation Research}, vol.~5, p. 100170, 2025.

\bibitem{liao2025cot}
H.~Liao, H.~Kong, B.~Wang, C.~Wang, W.~Ye, Z.~He, C.~Xu, and Z.~Li, ``Cot-drive: Efficient motion forecasting for autonomous driving with llms and chain-of-thought prompting,'' \emph{arXiv preprint arXiv:2503.07234}, 2025.

\bibitem{luo2025senserag}
X.~Luo, C.~Liu, F.~Ding, F.~Yang, Y.~Zhou, J.~Loo, and H.~H. Tew, ``Senserag: Constructing environmental knowledge bases with proactive querying for llm-based autonomous driving,'' in \emph{Proceedings of the Winter Conference on Applications of Computer Vision}, 2025, pp. 989--996.

\bibitem{mao2023gpt}
J.~Mao, Y.~Qian, J.~Ye, H.~Zhao, and Y.~Wang, ``Gpt-driver: Learning to drive with gpt,'' \emph{arXiv preprint arXiv:2310.01415}, 2023.

\bibitem{zheng2024planagent}
Y.~Zheng, Z.~Xing, Q.~Zhang, B.~Jin, P.~Li, Y.~Zheng, Z.~Xia, K.~Zhan, X.~Lang, Y.~Chen \emph{et~al.}, ``Planagent: A multi-modal large language agent for closed-loop vehicle motion planning,'' \emph{arXiv preprint arXiv:2406.01587}, 2024.

\bibitem{huang2024making}
Z.~Huang, T.~Tang, S.~Chen, S.~Lin, Z.~Jie, L.~Ma, G.~Wang, and X.~Liang, ``Making large language models better planners with reasoning-decision alignment,'' in \emph{European Conference on Computer Vision}.\hskip 1em plus 0.5em minus 0.4em\relax Springer, 2024, pp. 73--90.

\bibitem{yao2024calmm}
R.~Yao, Y.~Wang, H.~Liu, R.~Yang, Z.~Peng, L.~Zhu, and J.~Ma, ``Calmm-drive: Confidence-aware autonomous driving with large multimodal model,'' \emph{arXiv preprint arXiv:2412.04209}, 2024.

\bibitem{sha2310languagempc}
H.~Sha, Y.~Mu, Y.~Jiang, L.~Chen, C.~Xu, P.~Luo, S.~Li, M.~Tomizuka, W.~Zhan, and M.~Ding, ``Languagempc: Large language models as decision makers for autonomous driving. arxiv 2023,'' \emph{arXiv preprint arXiv:2310.03026}, 2023.

\bibitem{wang2023drivemlm}
W.~Wang, J.~Xie, C.~Hu, H.~Zou, J.~Fan, W.~Tong, Y.~Wen, S.~Wu, H.~Deng, Z.~Li \emph{et~al.}, ``Drivemlm: Aligning multi-modal large language models with behavioral planning states for autonomous driving,'' \emph{arXiv preprint arXiv:2312.09245}, 2023.

\bibitem{cui2024receive}
C.~Cui, Y.~Ma, X.~Cao, W.~Ye, and Z.~Wang, ``Receive, reason, and react: Drive as you say, with large language models in autonomous vehicles,'' \emph{IEEE Intelligent Transportation Systems Magazine}, 2024.

\bibitem{zhou2024safedrive}
Z.~Zhou, H.~Huang, B.~Li, S.~Zhao, Y.~Mu, and J.~Wang, ``Safedrive: Knowledge-and data-driven risk-sensitive decision-making for autonomous vehicles with large language models,'' \emph{arXiv preprint arXiv:2412.13238}, 2024.

\bibitem{jiang2024koma}
K.~Jiang, X.~Cai, Z.~Cui, A.~Li, Y.~Ren, H.~Yu, H.~Yang, D.~Fu, L.~Wen, and P.~Cai, ``Koma: Knowledge-driven multi-agent framework for autonomous driving with large language models,'' \emph{IEEE Transactions on Intelligent Vehicles}, 2024.

\bibitem{ma2025leapvad}
Y.~Ma, T.~Wei, N.~Zhong, J.~Mei, T.~Hu, L.~Wen, X.~Yang, B.~Shi, and Y.~Liu, ``Leapvad: A leap in autonomous driving via cognitive perception and dual-process thinking,'' \emph{arXiv preprint arXiv:2501.08168}, 2025.

\bibitem{fang2025towards}
S.~Fang, J.~Liu, M.~Ding, Y.~Cui, C.~Lv, P.~Hang, and J.~Sun, ``Towards interactive and learnable cooperative driving automation: a large language model-driven decision-making framework,'' \emph{IEEE Transactions on Vehicular Technology}, 2025.

\bibitem{fang2025interact}
S.~Fang, J.~Liu, C.~Xu, C.~Lv, P.~Hang, and J.~Sun, ``Interact, instruct to improve: A llm-driven parallel actor-reasoner framework for enhancing autonomous vehicle interactions,'' \emph{arXiv preprint arXiv:2503.00502}, 2025.

\bibitem{ren2025cot}
T.~Ren, H.~Hu, J.~Zuo, X.~Chen, J.~Wang, C.~J. Xue, J.-M. Wu, and N.~Guan, ``Cot-vlm4tar: Chain-of-thought guided vision-language models for traffic anomaly resolution,'' \emph{arXiv preprint arXiv:2503.01632}, 2025.

\bibitem{chen2024driving}
L.~Chen, O.~Sinavski, J.~H{\"u}nermann, A.~Karnsund, A.~J. Willmott, D.~Birch, D.~Maund, and J.~Shotton, ``Driving with llms: Fusing object-level vector modality for explainable autonomous driving,'' in \emph{2024 IEEE International Conference on Robotics and Automation (ICRA)}.\hskip 1em plus 0.5em minus 0.4em\relax IEEE, 2024, pp. 14\,093--14\,100.

\bibitem{luo2024pkrd}
X.~Luo, F.~Ding, Y.~Song, X.~Zhang, and J.~Loo, ``Pkrd-cot: A unified chain-of-thought prompting for multi-modal large language models in autonomous driving,'' \emph{arXiv preprint arXiv:2412.02025}, 2024.

\bibitem{shao2024lmdrive}
H.~Shao, Y.~Hu, L.~Wang, G.~Song, S.~L. Waslander, Y.~Liu, and H.~Li, ``Lmdrive: Closed-loop end-to-end driving with large language models,'' in \emph{Proceedings of the IEEE/CVF Conference on Computer Vision and Pattern Recognition}, 2024, pp. 15\,120--15\,130.

\bibitem{jiang2024senna}
B.~Jiang, S.~Chen, B.~Liao, X.~Zhang, W.~Yin, Q.~Zhang, C.~Huang, W.~Liu, and X.~Wang, ``Senna: Bridging large vision-language models and end-to-end autonomous driving,'' \emph{arXiv preprint arXiv:2410.22313}, 2024.

\bibitem{hwang2024emma}
J.-J. Hwang, R.~Xu, H.~Lin, W.-C. Hung, J.~Ji, K.~Choi, D.~Huang, T.~He, P.~Covington, B.~Sapp \emph{et~al.}, ``Emma: End-to-end multimodal model for autonomous driving,'' \emph{arXiv preprint arXiv:2410.23262}, 2024.

\bibitem{xing2025openemma}
S.~Xing, C.~Qian, Y.~Wang, H.~Hua, K.~Tian, Y.~Zhou, and Z.~Tu, ``Openemma: Open-source multimodal model for end-to-end autonomous driving,'' in \emph{Proceedings of the Winter Conference on Applications of Computer Vision}, 2025, pp. 1001--1009.

\bibitem{qiao2025lightemma}
Z.~Qiao, H.~Li, Z.~Cao, and H.~X. Liu, ``Lightemma: Lightweight end-to-end multimodal model for autonomous driving,'' \emph{arXiv preprint arXiv:2505.00284}, 2025.

\bibitem{fu2025orion}
H.~Fu, D.~Zhang, Z.~Zhao, J.~Cui, D.~Liang, C.~Zhang, D.~Zhang, H.~Xie, B.~Wang, and X.~Bai, ``Orion: A holistic end-to-end autonomous driving framework by vision-language instructed action generation,'' \emph{arXiv preprint arXiv:2503.19755}, 2025.

\bibitem{sima2024drivelm}
C.~Sima, K.~Renz, K.~Chitta, L.~Chen, H.~Zhang, C.~Xie, J.~Bei{\ss}wenger, P.~Luo, A.~Geiger, and H.~Li, ``Drivelm: Driving with graph visual question answering,'' in \emph{European Conference on Computer Vision}.\hskip 1em plus 0.5em minus 0.4em\relax Springer, 2024, pp. 256--274.

\bibitem{zhao2025sce2drivex}
R.~Zhao, Q.~Yuan, J.~Li, H.~Hu, Y.~Li, C.~Zheng, and F.~Gao, ``Sce2drivex: A generalized mllm framework for scene-to-drive learning,'' \emph{arXiv preprint arXiv:2502.14917}, 2025.

\bibitem{mandalika2025primedrive}
S.~Mandalika, A.~Nambiar \emph{et~al.}, ``Primedrive-cot: A precognitive chain-of-thought framework for uncertainty-aware object interaction in driving scene scenario,'' \emph{arXiv preprint arXiv:2504.05908}, 2025.

\bibitem{gao2025langcoop}
X.~Gao, Y.~Wu, R.~Wang, C.~Liu, Y.~Zhou, and Z.~Tu, ``Langcoop: Collaborative driving with language,'' \emph{arXiv preprint arXiv:2504.13406}, 2025.

\bibitem{liu2025x}
W.~Liu, J.~Zhang, B.~Zheng, Y.~Hu, Y.~Lin, and Z.~Zeng, ``X-driver: Explainable autonomous driving with vision-language models,'' \emph{arXiv preprint arXiv:2505.05098}, 2025.

\bibitem{zhou2022least}
D.~Zhou, N.~Sch{\"a}rli, L.~Hou, J.~Wei, N.~Scales, X.~Wang, D.~Schuurmans, C.~Cui, O.~Bousquet, Q.~Le \emph{et~al.}, ``Least-to-most prompting enables complex reasoning in large language models,'' \emph{arXiv preprint arXiv:2205.10625}, 2022.

\bibitem{bai2023qwen}
J.~Bai, S.~Bai, Y.~Chu, Z.~Cui, K.~Dang, X.~Deng, Y.~Fan, W.~Ge, Y.~Han, F.~Huang \emph{et~al.}, ``Qwen technical report,'' \emph{arXiv preprint arXiv:2309.16609}, 2023.

\bibitem{jiang2024survey}
J.~Jiang, F.~Wang, J.~Shen, S.~Kim, and S.~Kim, ``A survey on large language models for code generation,'' \emph{arXiv preprint arXiv:2406.00515}, 2024.

\bibitem{kouvaritakis2016model}
B.~Kouvaritakis and M.~Cannon, ``Model predictive control,'' \emph{Switzerland: Springer International Publishing}, vol.~38, no. 13-56, p.~7, 2016.

\bibitem{evans1984heuristic}
J.~S.~B. Evans, ``Heuristic and analytic processes in reasoning,'' \emph{British Journal of Psychology}, vol.~75, no.~4, pp. 451--468, 1984.

\bibitem{li2025system}
Z.-Z. Li, D.~Zhang, M.-L. Zhang, J.~Zhang, Z.~Liu, Y.~Yao, H.~Xu, J.~Zheng, P.-J. Wang, X.~Chen \emph{et~al.}, ``From system 1 to system 2: A survey of reasoning large language models,'' \emph{arXiv preprint arXiv:2502.17419}, 2025.

\bibitem{liu2021survey}
W.~Liu, Q.~Dong, P.~Wang, G.~Yang, L.~Meng, Y.~Song, Y.~Shi, and Y.~Xue, ``A survey on autonomous driving datasets,'' in \emph{2021 8th International Conference on Dependable Systems and Their Applications (DSA)}.\hskip 1em plus 0.5em minus 0.4em\relax IEEE, 2021, pp. 399--407.

\bibitem{liu2024survey}
M.~Liu, E.~Yurtsever, J.~Fossaert, X.~Zhou, W.~Zimmer, Y.~Cui, B.~L. Zagar, and A.~C. Knoll, ``A survey on autonomous driving datasets: Statistics, annotation quality, and a future outlook,'' \emph{IEEE Transactions on Intelligent Vehicles}, 2024.

\bibitem{li2019aads}
W.~Li, C.~Pan, R.~Zhang, J.~Ren, Y.~Ma, J.~Fang, F.~Yan, Q.~Geng, X.~Huang, H.~Gong \emph{et~al.}, ``Aads: Augmented autonomous driving simulation using data-driven algorithms,'' \emph{Science robotics}, vol.~4, no.~28, p. eaaw0863, 2019.

\bibitem{xie2025vlms}
S.~Xie, L.~Kong, Y.~Dong, C.~Sima, W.~Zhang, Q.~A. Chen, Z.~Liu, and L.~Pan, ``Are vlms ready for autonomous driving? an empirical study from the reliability, data, and metric perspectives,'' \emph{arXiv preprint arXiv:2501.04003}, 2025.

\bibitem{deruyttere2019talk2car}
T.~Deruyttere, S.~Vandenhende, D.~Grujicic, L.~Van~Gool, and M.-F. Moens, ``Talk2car: Taking control of your self-driving car,'' \emph{arXiv preprint arXiv:1909.10838}, 2019.

\bibitem{qian2024nuscenes}
T.~Qian, J.~Chen, L.~Zhuo, Y.~Jiao, and Y.-G. Jiang, ``Nuscenes-qa: A multi-modal visual question answering benchmark for autonomous driving scenario,'' in \emph{Proceedings of the AAAI Conference on Artificial Intelligence}, vol.~38, no.~5, 2024, pp. 4542--4550.

\bibitem{choudhary2024talk2bev}
T.~Choudhary, V.~Dewangan, S.~Chandhok, S.~Priyadarshan, A.~Jain, A.~K. Singh, S.~Srivastava, K.~M. Jatavallabhula, and K.~M. Krishna, ``Talk2bev: Language-enhanced bird’s-eye view maps for autonomous driving,'' in \emph{2024 IEEE International Conference on Robotics and Automation (ICRA)}.\hskip 1em plus 0.5em minus 0.4em\relax IEEE, 2024, pp. 16\,345--16\,352.

\bibitem{inoue2024nuscenes}
Y.~Inoue, Y.~Yada, K.~Tanahashi, and Y.~Yamaguchi, ``Nuscenes-mqa: Integrated evaluation of captions and qa for autonomous driving datasets using markup annotations,'' in \emph{Proceedings of the IEEE/CVF Winter Conference on Applications of Computer Vision}, 2024, pp. 930--938.

\bibitem{guo2024drivemllm}
X.~Guo, R.~Zhang, Y.~Duan, Y.~He, C.~Zhang, S.~Liu, and L.~Chen, ``Drivemllm: A benchmark for spatial understanding with multimodal large language models in autonomous driving,'' \emph{arXiv preprint arXiv:2411.13112}, 2024.

\bibitem{wu2025language}
D.~Wu, W.~Han, Y.~Liu, T.~Wang, C.-z. Xu, X.~Zhang, and J.~Shen, ``Language prompt for autonomous driving,'' in \emph{Proceedings of the AAAI Conference on Artificial Intelligence}, vol.~39, no.~8, 2025, pp. 8359--8367.

\bibitem{ishaq2025drivelmm}
A.~Ishaq, J.~Lahoud, K.~More, O.~Thawakar, R.~Thawkar, D.~Dissanayake, N.~Ahsan, Y.~Li, F.~S. Khan, H.~Cholakkal \emph{et~al.}, ``Drivelmm-o1: A step-by-step reasoning dataset and large multimodal model for driving scenario understanding,'' \emph{arXiv preprint arXiv:2503.10621}, 2025.

\bibitem{xu2020explainable}
Y.~Xu, X.~Yang, L.~Gong, H.-C. Lin, T.-Y. Wu, Y.~Li, and N.~Vasconcelos, ``Explainable object-induced action decision for autonomous vehicles,'' in \emph{Proceedings of the IEEE/CVF Conference on Computer Vision and Pattern Recognition}, 2020, pp. 9523--9532.

\bibitem{kim2018textual}
J.~Kim, A.~Rohrbach, T.~Darrell, J.~Canny, and Z.~Akata, ``Textual explanations for self-driving vehicles,'' in \emph{Proceedings of the European conference on computer vision (ECCV)}, 2018, pp. 563--578.

\bibitem{xu2024drivegpt4}
Z.~Xu, Y.~Zhang, E.~Xie, Z.~Zhao, Y.~Guo, K.-Y.~K. Wong, Z.~Li, and H.~Zhao, ``Drivegpt4: Interpretable end-to-end autonomous driving via large language model,'' \emph{IEEE Robotics and Automation Letters}, 2024.

\bibitem{wu2023referring}
D.~Wu, W.~Han, T.~Wang, X.~Dong, X.~Zhang, and J.~Shen, ``Referring multi-object tracking,'' in \emph{Proceedings of the IEEE/CVF conference on computer vision and pattern recognition}, 2023, pp. 14\,633--14\,642.

\bibitem{feng2021cityflow}
Q.~Feng, V.~Ablavsky, and S.~Sclaroff, ``Cityflow-nl: Tracking and retrieval of vehicles at city scale by natural language descriptions,'' \emph{arXiv preprint arXiv:2101.04741}, 2021.

\bibitem{corbiere2025drivingvqa}
C.~Corbi{\`e}re, S.~Roburin, S.~Montariol, A.~Bosselut, and A.~Alahi, ``Drivingvqa: Analyzing visual chain-of-thought reasoning of vision language models in real-world scenarios with driving theory tests,'' \emph{arXiv preprint arXiv:2501.04671}, 2025.

\bibitem{cao2024maplm}
X.~Cao, T.~Zhou, Y.~Ma, W.~Ye, C.~Cui, K.~Tang, Z.~Cao, K.~Liang, Z.~Wang, J.~M. Rehg \emph{et~al.}, ``Maplm: A real-world large-scale vision-language benchmark for map and traffic scene understanding,'' in \emph{Proceedings of the IEEE/CVF Conference on Computer Vision and Pattern Recognition}, 2024, pp. 21\,819--21\,830.

\bibitem{sachdeva2024rank2tell}
E.~Sachdeva, N.~Agarwal, S.~Chundi, S.~Roelofs, J.~Li, M.~Kochenderfer, C.~Choi, and B.~Dariush, ``Rank2tell: A multimodal driving dataset for joint importance ranking and reasoning,'' in \emph{Proceedings of the IEEE/CVF winter conference on applications of computer vision}, 2024, pp. 7513--7522.

\bibitem{malla2023drama}
S.~Malla, C.~Choi, I.~Dwivedi, J.~H. Choi, and J.~Li, ``Drama: Joint risk localization and captioning in driving,'' in \emph{Proceedings of the IEEE/CVF winter conference on applications of computer vision}, 2023, pp. 1043--1052.

\bibitem{marcu2024lingoqa}
A.-M. Marcu, L.~Chen, J.~H{\"u}nermann, A.~Karnsund, B.~Hanotte, P.~Chidananda, S.~Nair, V.~Badrinarayanan, A.~Kendall, J.~Shotton \emph{et~al.}, ``Lingoqa: Visual question answering for autonomous driving,'' in \emph{European Conference on Computer Vision}.\hskip 1em plus 0.5em minus 0.4em\relax Springer, 2024, pp. 252--269.

\bibitem{geiger2013vision}
A.~Geiger, P.~Lenz, C.~Stiller, and R.~Urtasun, ``Vision meets robotics: The kitti dataset,'' \emph{The international journal of robotics research}, vol.~32, no.~11, pp. 1231--1237, 2013.

\bibitem{yu2020bdd100k}
F.~Yu, H.~Chen, X.~Wang, W.~Xian, Y.~Chen, F.~Liu, V.~Madhavan, and T.~Darrell, ``Bdd100k: A diverse driving dataset for heterogeneous multitask learning,'' in \emph{Proceedings of the IEEE/CVF conference on computer vision and pattern recognition}, 2020, pp. 2636--2645.

\bibitem{caesar2020nuscenes}
H.~Caesar, V.~Bankiti, A.~H. Lang, S.~Vora, V.~E. Liong, Q.~Xu, A.~Krishnan, Y.~Pan, G.~Baldan, and O.~Beijbom, ``nuscenes: A multimodal dataset for autonomous driving,'' in \emph{Proceedings of the IEEE/CVF conference on computer vision and pattern recognition}, 2020, pp. 11\,621--11\,631.

\bibitem{sun2020scalability}
P.~Sun, H.~Kretzschmar, X.~Dotiwalla, A.~Chouard, V.~Patnaik, P.~Tsui, J.~Guo, Y.~Zhou, Y.~Chai, B.~Caine \emph{et~al.}, ``Scalability in perception for autonomous driving: Waymo open dataset,'' in \emph{Proceedings of the IEEE/CVF conference on computer vision and pattern recognition}, 2020, pp. 2446--2454.

\bibitem{krajewski2018highd}
R.~Krajewski, J.~Bock, L.~Kloeker, and L.~Eckstein, ``The highd dataset: A drone dataset of naturalistic vehicle trajectories on german highways for validation of highly automated driving systems,'' in \emph{2018 21st international conference on intelligent transportation systems (ITSC)}.\hskip 1em plus 0.5em minus 0.4em\relax IEEE, 2018, pp. 2118--2125.

\bibitem{coifman2017critical}
B.~Coifman and L.~Li, ``A critical evaluation of the next generation simulation (ngsim) vehicle trajectory dataset,'' \emph{Transportation Research Part B: Methodological}, vol. 105, pp. 362--377, 2017.

\bibitem{tang2019cityflow}
Z.~Tang, M.~Naphade, M.-Y. Liu, X.~Yang, S.~Birchfield, S.~Wang, R.~Kumar, D.~Anastasiu, and J.-N. Hwang, ``Cityflow: A city-scale benchmark for multi-target multi-camera vehicle tracking and re-identification,'' in \emph{Proceedings of the IEEE/CVF conference on computer vision and pattern recognition}, 2019, pp. 8797--8806.

\bibitem{li2025explainable}
M.~Li, Z.~Cui, Y.~Wang, Y.~Huang, and H.~Chen, ``An explainable $ q $-learning method for longitudinal control of autonomous vehicles,'' \emph{IEEE Transactions on Intelligent Transportation Systems}, 2025.

\bibitem{lobo2024impact}
E.~Lobo, C.~Agarwal, and H.~Lakkaraju, ``On the impact of fine-tuning on chain-of-thought reasoning,'' \emph{arXiv preprint arXiv:2411.15382}, 2024.

\bibitem{atakishiyev2023explaining}
S.~Atakishiyev, M.~Salameh, H.~Babiker, and R.~Goebel, ``Explaining autonomous driving actions with visual question answering,'' in \emph{2023 IEEE 26th International Conference on Intelligent Transportation Systems (ITSC)}.\hskip 1em plus 0.5em minus 0.4em\relax IEEE, 2023, pp. 1207--1214.

\bibitem{mao2021one}
J.~Mao, M.~Niu, C.~Jiang, H.~Liang, J.~Chen, X.~Liang, Y.~Li, C.~Ye, W.~Zhang, Z.~Li \emph{et~al.}, ``One million scenes for autonomous driving: Once dataset,'' \emph{arXiv preprint arXiv:2106.11037}, 2021.

\bibitem{dosovitskiy2017carla}
A.~Dosovitskiy, G.~Ros, F.~Codevilla, A.~Lopez, and V.~Koltun, ``Carla: An open urban driving simulator,'' in \emph{Conference on robot learning}.\hskip 1em plus 0.5em minus 0.4em\relax PMLR, 2017, pp. 1--16.

\bibitem{song2023synthetic}
Z.~Song, Z.~He, X.~Li, Q.~Ma, R.~Ming, Z.~Mao, H.~Pei, L.~Peng, J.~Hu, D.~Yao \emph{et~al.}, ``Synthetic datasets for autonomous driving: A survey,'' \emph{IEEE Transactions on Intelligent Vehicles}, vol.~9, no.~1, pp. 1847--1864, 2023.

\bibitem{li2024think2drive}
Q.~Li, X.~Jia, S.~Wang, and J.~Yan, ``Think2drive: Efficient reinforcement learning by thinking with latent world model for autonomous driving (in carla-v2),'' in \emph{European Conference on Computer Vision}.\hskip 1em plus 0.5em minus 0.4em\relax Springer, 2024, pp. 142--158.

\bibitem{guo2019safe}
J.~Guo, U.~Kurup, and M.~Shah, ``Is it safe to drive? an overview of factors, metrics, and datasets for driveability assessment in autonomous driving,'' \emph{IEEE Transactions on Intelligent Transportation Systems}, vol.~21, no.~8, pp. 3135--3151, 2019.

\bibitem{golovneva2022roscoe}
O.~Golovneva, M.~Chen, S.~Poff, M.~Corredor, L.~Zettlemoyer, M.~Fazel-Zarandi, and A.~Celikyilmaz, ``Roscoe: A suite of metrics for scoring step-by-step reasoning,'' \emph{arXiv preprint arXiv:2212.07919}, 2022.

\bibitem{papineni2002bleu}
K.~Papineni, S.~Roukos, T.~Ward, and W.-J. Zhu, ``Bleu: a method for automatic evaluation of machine translation,'' in \emph{Proceedings of the 40th annual meeting of the Association for Computational Linguistics}, 2002, pp. 311--318.

\bibitem{vedantam2015cider}
R.~Vedantam, C.~Lawrence~Zitnick, and D.~Parikh, ``Cider: Consensus-based image description evaluation,'' in \emph{Proceedings of the IEEE conference on computer vision and pattern recognition}, 2015, pp. 4566--4575.

\bibitem{banerjee2005meteor}
S.~Banerjee and A.~Lavie, ``Meteor: An automatic metric for mt evaluation with improved correlation with human judgments,'' in \emph{Proceedings of the acl workshop on intrinsic and extrinsic evaluation measures for machine translation and/or summarization}, 2005, pp. 65--72.

\bibitem{liu2023visual}
H.~Liu, C.~Li, Q.~Wu, and Y.~J. Lee, ``Visual instruction tuning,'' \emph{Advances in neural information processing systems}, vol.~36, pp. 34\,892--34\,916, 2023.

\bibitem{xing2024autotrust}
S.~Xing, H.~Hua, X.~Gao, S.~Zhu, R.~Li, K.~Tian, X.~Li, H.~Huang, T.~Yang, Z.~Wang \emph{et~al.}, ``Autotrust: Benchmarking trustworthiness in large vision language models for autonomous driving,'' \emph{arXiv preprint arXiv:2412.15206}, 2024.

\bibitem{kim2024openvla}
M.~J. Kim, K.~Pertsch, S.~Karamcheti, T.~Xiao, A.~Balakrishna, S.~Nair, R.~Rafailov, E.~Foster, G.~Lam, P.~Sanketi \emph{et~al.}, ``Openvla: An open-source vision-language-action model,'' \emph{arXiv preprint arXiv:2406.09246}, 2024.

\bibitem{li2024large}
Y.~Li, K.~Katsumata, E.~Javanmardi, and M.~Tsukada, ``Large language models for human-like autonomous driving: A survey,'' in \emph{2024 IEEE 27th International Conference on Intelligent Transportation Systems (ITSC)}.\hskip 1em plus 0.5em minus 0.4em\relax IEEE, 2024, pp. 439--446.

\bibitem{song2024preference}
F.~Song, B.~Yu, M.~Li, H.~Yu, F.~Huang, Y.~Li, and H.~Wang, ``Preference ranking optimization for human alignment,'' in \emph{Proceedings of the AAAI Conference on Artificial Intelligence}, vol.~38, no.~17, 2024, pp. 18\,990--18\,998.

\bibitem{kothawade2021auto}
S.~Kothawade, V.~Khandelwal, K.~Basu, H.~Wang, and G.~Gupta, ``Auto-discern: autonomous driving using common sense reasoning,'' \emph{arXiv preprint arXiv:2110.13606}, 2021.

\bibitem{xu2025chain}
S.~Xu, W.~Xie, L.~Zhao, and P.~He, ``Chain of draft: Thinking faster by writing less,'' \emph{arXiv preprint arXiv:2502.18600}, 2025.

\bibitem{ji2023survey}
Z.~Ji, N.~Lee, R.~Frieske, T.~Yu, D.~Su, Y.~Xu, E.~Ishii, Y.~J. Bang, A.~Madotto, and P.~Fung, ``Survey of hallucination in natural language generation,'' \emph{ACM computing surveys}, vol.~55, no.~12, pp. 1--38, 2023.

\bibitem{deng2024explicit}
Y.~Deng, Y.~Choi, and S.~Shieber, ``From explicit cot to implicit cot: Learning to internalize cot step by step,'' \emph{arXiv preprint arXiv:2405.14838}, 2024.

\bibitem{kong2025scalable}
D.~Kong, M.~Zhao, D.~Xu, B.~Pang, S.~Wang, E.~Honig, Z.~Si, C.~Li, J.~Xie, S.~Xie \emph{et~al.}, ``Scalable language models with posterior inference of latent thought vectors,'' \emph{arXiv preprint arXiv:2502.01567}, 2025.

\end{thebibliography}



\end{document}